\documentclass{article}
\PassOptionsToPackage{table}{xcolor}

\usepackage[numbers]{natbib}
\usepackage[preprint]{main}
\usepackage{amsmath, amssymb}
\usepackage{graphicx}
\usepackage{booktabs}
\usepackage{hyperref}
\usepackage{subcaption}
\usepackage{enumitem}

\usepackage{xspace}
\usepackage{adjustbox}
\usepackage{array}
\usepackage{float}
\usepackage{geometry}
\usepackage{dblfloatfix}
\usepackage{setspace}
\usepackage{footmisc}
\usepackage{multirow}
\usepackage{longtable}
\usepackage{tabularx}
\usepackage{xltabular}
\usepackage{threeparttable}
\usepackage{siunitx}
\usepackage{fancyvrb}    %
\usepackage[most]{tcolorbox}  %

\usepackage{amsfonts}
\usepackage{mathtools}
\usepackage{bm}
\usepackage{dsfont}

\usepackage{tikz}
\usepackage{pgfplots}
\pgfplotsset{compat=1.18}
\usepackage{algorithm}
\usepackage{algpseudocode}
\usepackage{listings}

\usepackage{cleveref}
\hypersetup{colorlinks=true}

\usepackage{todonotes}
\usepackage{fontawesome5}
\usepackage{CJKutf8}
\usepackage{lipsum}
\usepackage{xcolor}

\title{Chain-of-Experts: Unlocking the Communication Power of Mixture-of-Experts Models}

\author{
\makebox[\textwidth][c]{%
\textbf{Zihan Wang\textsuperscript{1*}}, 
\textbf{Rui Pan\textsuperscript{2*}}, 
Jiarui Yao\textsuperscript{2}, 
Róbert Csordás\textsuperscript{3}, 
Linjie Li\textsuperscript{4}, 
Lu Yin\textsuperscript{5}}\\
\makebox[\textwidth][c]{%
\textbf{Jiajun Wu\textsuperscript{3}, 
Tong Zhang\textsuperscript{2}, 
Manling Li\textsuperscript{1$\dagger$}, 
Shiwei Liu\textsuperscript{6$\dagger$}}} \\[0.2cm]
\makebox[\textwidth][c]{%
\textsuperscript{1}Northwestern University \quad
\textsuperscript{2}University of Illinois Urbana-Champaign \quad
\textsuperscript{3}Stanford University}\\
\makebox[\textwidth][c]{%
\textsuperscript{4}University of Washington \quad
\textsuperscript{5}University of Surrey \quad
\textsuperscript{6}University of Oxford}
}

\makeatletter
\let\@oldmaketitle\@maketitle
\def\@maketitle{%
  \@oldmaketitle
  \footnotetext[1]{Equal contribution.}
  \footnotetext[2]{Equal advising.}
}
\makeatother

\begin{document}

\maketitle

\begin{abstract}

We propose \textbf{Chain-of-Experts (CoE)}, a new Mixture-of-Experts (MoE) architecture that introduces sequential expert communication within each layer. Unlike traditional MoE models, where experts operate independently in parallel, CoE processes tokens iterative across a chain of experts inside a layer. To support dynamic expert selection across iterations, CoE employs a dedicated router at each iteration step within a layer. This design allows tokens to re-evaluate and select different experts during each iteration, rather than being statically assigned. As a result, CoE introduces a flexible routing mechanism that increases the diversity of expert combinations and enriches the model’s representational capacity. 
CoE demonstrates improved performance under fixed compute: on Math reasoning tasks, it reduces validation loss from 1.20 to 1.12 compared to a standard MoE. Beyond performance, CoE offers a new \textit{scaling axis}—depth through expert iteration—which complements conventional width/depth scaling. For example, using 2× iterations matches the performance of 3× expert selections (in width), while reducing memory usage by 17.6–42\% relative to other scaling strategies.
Our analysis reveals that CoE’s benefits stem from its iterative residual structure and enhanced expert specialization empowered by iterative routing, which together unlock more expressive representations. Code is available at  \href{https://github.com/ZihanWang314/coe}{\texttt{https://github.com/ZihanWang314/coe}}.

\end{abstract}

\section{Introduction}

Mixture-of-Experts (MoE) architectures have emerged as a dominant strategy for scaling large language models (LLMs), delivering significant gains in compute efficiency by sparsely activating a small subset of expert networks per input token~\citep{shazeer2017outrageously, lepikhin2020gshard,fedus2022switch, zoph2022designing,zhou2022moe}. This sparsity enables models to scale their parameter count independently from runtime cost, allowing for improved memory utilization and throughput during inference. Recent MoE systems have demonstrated strong empirical performance across a wide range of domains—including language~\citep{deepseekmoe, abdin2024phi3, meta2025llama4, qwen2025qwen3}, reasoning~\citep{jiang2024mixtral, muennighoff2024olmoe}, and multimodal vision-language tasks~\citep{riquelme2021scalingvisionsparsemixture, li2024unimoescalingunifiedmultimodal}.

Progress in MoE research has largely centered on scaling up expert capacity~\citep{tan2023sparse, csordas2024moeut}, improving routing algorithms~\citep{deepseekmoe, qiu2024layerwise, zhou2022moe}, and enhancing training stability~\citep{riquelme2021scalingvisionsparsemixture}. These approaches often rely on a shared architectural assumption: \emph{experts are conditionally and independently activated in parallel}, with no explicit interaction between them. While this design maximizes parallelism and system efficiency, it may also constrain the model’s ability to exploit complementary reasoning patterns across experts. As a result, existing MoE models could underutilize their available capacity, particularly for complex tasks that benefit from multi-expert coordination.

\begin{figure}[t]
    \centering
    \includegraphics[width=\linewidth]{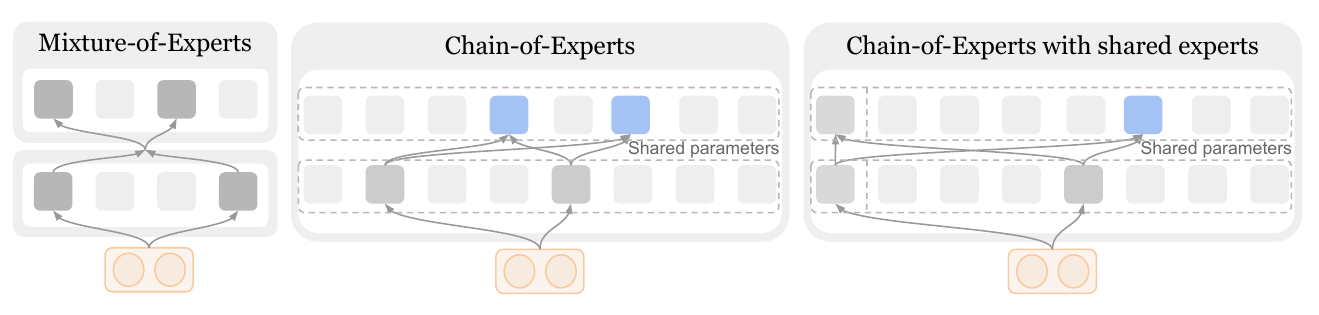}
    \caption{\textbf{Comparison between Mixture-of-Experts (MoE) and Chain-of-Experts (CoE).} Under the same model depth and parameters, CoE enables iterative expert communication to offer more flexible expert choice compared to MoE.}
    \label{fig:samebudget}
\end{figure}

We challenge this independence assumption by introducing \textbf{Chain-of-Experts (CoE)} (Figure~\ref{fig:samebudget}), a new MoE framework that enables \emph{sequential communication} among intra-layer experts through iterative processing. While keeping the total number of experts used to process a token within a layer unchanged, CoE allows experts to process tokens in a ``relay race" manner rather than independently: experts receive the intermediate representation from their predecessors, process it, and pass it to their successors, enabling richer expert composition and higher effective depth. Furthermore, sequential expert chains introduces a new scaling axis—\emph{depth through iteration}—which could complement or even surpasses traditional scaling via width or layer depth. 

In our experiments, under equivalent compute, CoE yields better performance (e.g., math validation loss drops from 1.20 to 1.12), higher efficiency (17.6–42\% lower memory usage with comparable accuracy), and greater specialization (823$\times$ more effective expert combinations). These gains stem from a simple change: \textbf{iterative MoE layers with independent routing and residual connections at each iteration}. Our design is grounded in prior observations that experts often learn complementary roles and are highly specialized for different uses~\cite{wang2024letexpertsticklast}. We empirically show that CoE consistently outperforms standard MoEs on reasoning-intensive tasks, particularly under compute-constrained regimes. We further analyze the impact of iteration count, gating design, and communication depth, offering a new lens on scaling modular language models efficiently.

\section{Background: Mixture-of-Experts Transformers}

We begin by formalizing the output computation of a standard Mixture-of-Experts (MoE) layer, a widely adopted mechanism to increase model capacity without a proportional increase in computation cost. In a typical Transformer architecture, certain Feed-Forward Network (FFN) sublayers are replaced by sparse MoE modules, where only a small subset of the available experts are activated per token.

Let $x \in \mathbb{R}^d$ denote the input token embedding, and let $\{E_i(\cdot)\}_{i=1}^N$ be a set of $N$ expert networks, each structurally identical to a standard FFN. A gating function determines which $K \ll N$ experts are selected for a given input, assigning a nonzero routing weight $g_i$ to each selected expert. The output of the MoE layer is computed as:
\begin{equation}
y = \sum_{i=1}^N g_i \cdot E_i(x), \label{eq:moe_output}
\end{equation}
where the gating weights $g_i$ are defined as:
\begin{equation}
g_i = 
\begin{cases}
s_i, & s_i \in \text{TopK}(\{s_j\}_{j=1}^N, K), \\
0,   & \text{otherwise},
\end{cases}
\quad \text{with} \quad
s_i = \text{Softmax}(e_i^\top x). \label{eq:gating}
\end{equation}
Here, $e_i \in \mathbb{R}^d$ is the router vector associated with the $i$-th expert. The dot product $e_i^\top x$ measures the affinity between the input token and expert $E_i$. The $\text{TopK}$ operator selects the $K$ experts with the highest affinity scores, and the $\text{Softmax}$ normalizes these scores over all experts. This gating mechanism ensures sparsity: only $K$ experts are active per token, significantly reducing compute and memory usage.

Some recent MoE variants \citep{dai2024deepseekmoe} introduce \textit{shared experts} across all tokens or layers to improve generalization and parameter efficiency. In such cases, the MoE output is extended as follows:

\begin{equation}
y = \sum_{i=1}^{M} \hat{E}_i(x) + \sum_{i=1}^{N} g_i \cdot E_i(x), \label{eq:shared_experts}
\end{equation}

where $\{\hat{E}_i(\cdot)\}_{i=1}^M$ denotes a set of shared experts that are applied uniformly to all tokens, while $\{E_i(\cdot)\}_{i=1}^N$ are sparsely gated token-specific experts. Together, these mechanisms define the standard MoE formulation used in many scalable Transformer-based language models~\citep{shazeer2017outrageously, fedus2022switch, zoph2022designing, deepseekmoe}. Our work builds on top of this foundation but departs from the conventional \emph{parallel and independent} expert structure by introducing \emph{sequential communication} across selected experts.

\begin{figure}[t]
    \centering
    \includegraphics[width=\linewidth]{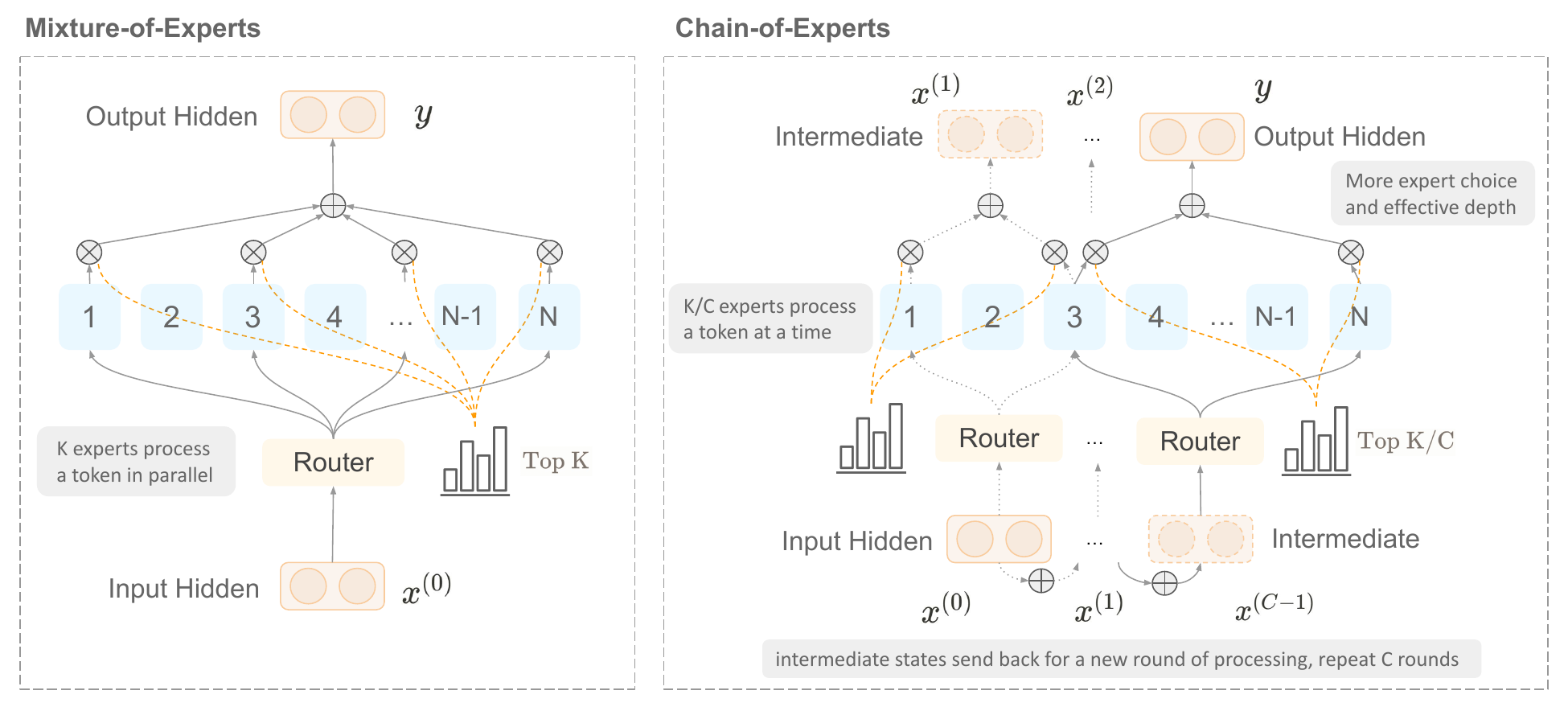}
    \caption{
    \textbf{Illustration of Chain-of-Experts. }
    MoE Top-K experts operate in parallel without interaction, CoE enables the same number of experts to process sequentially via intermediate representations, allowing deeper network processing with the same per-token expert processing. Residual connections are enabled and iteration-independent routers are used for more effective training.
    }
    \label{fig:main}
\end{figure}

\section{Chain-of-Experts: Communicative Expert Processing}

While standard MoE routes each token through a fixed set of experts in a single forward pass, this design inhibits communication across experts within each layer. Intuitively, a token might benefit from passing through multiple experts in sequence—allowing each to refine the representation conditioned on what others have already processed. 

To capture this notion of iterative processing, we propose an expert chaining mechanism that processes tokens over $C$ \textbf{communication steps}, named Chain-of-Experts (CoE), as an alternative of current MoE layers. As shown in Figure~\ref{fig:main}, different from MoE layers, the token is re-routed to a new set of experts, each seeing the hidden state produced in the previous step. 
Formally, given an input token embedding $x$, we initialize the first hidden state as:
\begin{equation}
x^{(0)} = x,
\end{equation}
Then, at each iteration $t = 1, \dots, C$, we compute the next hidden state as:
\begin{equation}
x^{(t)} = \sum_{i=1}^{N} g_{t,i} \cdot E_i(x^{(t-1)}) + \mathbb{I}_r \cdot x^{(t-1)},
\end{equation}
where $E_i$ is the $i$-th expert and $g_{t,i}$ is the gating weight at step $t$. The term $\mathbb{I}_r \cdot x^{(t-1)}$ denotes a residual connection, which we include by setting $\mathbb{I}_r = 1$. This simple addition helps preserve information and stabilizes updates across iterations.
The output after $C$ iterations is:
\begin{equation}
y = x^{(C)}.
\end{equation}
This formulation naturally introduces expert communication—each expert in round $t$ builds on what was computed in round $t{-}1$. Instead of forcing all experts to work independently, the model learns to decompose reasoning across steps. To ensure computational cost remains comparable to standard MoE, we could select only $K/C$ experts at each iteration. 

\vspace{0.5em}
\noindent\textbf{Iteration-based Independent Routing.}
To enable tokens to reevaluate and make adaptive routing decisions based on progressively refined hidden states, we assign a separate router to each iteration. Specifically, the gating weight $g_{t,i}$ is defined as:
\begin{equation}
g_{t,i} = 
\begin{cases}
s_{t,i}, & s_{t,i} \in \text{TopK}(s_{t,j} \mid 1 \leq j \leq N, K/C) \\
0, & \text{otherwise}
\end{cases},
\quad
\end{equation}
where $s_{t,i} = \text{Softmax}(e_{t,i}^\top x^{(t-1)})$.
This independent routing mechanism across iterations allows each step to adapt freely: experts engaged in later iterations can refine, reinterpret, or build upon the intermediate representations produced earlier, enabling a form of progressive, layered processing.

Our iterative framework transforms MoE from a shallowly parallel system into a sequential expert reasoning process, while preserving its sparsity and modularity.
When aligning both depth and total expert parameters, CoE offers a richer set of expert combinations compared to standard MoE by enabling expert reuse. As illustrated in Figure~\ref{fig:samebudget}, we compare a two-layer Transformer equipped with MoE, where each layer contains four experts and activates two per token. With CoE, under the same parameter budget, we can achieve comparable depth using only a single Transformer layer with two communication steps. This setup allows more diverse expert compositions without significantly increasing computational cost.
It is worth noting that Universal Transformers~\cite{Dehghani2019Universal}, such as MoEUT~\cite{csordas2024moeut}, also adopt a form of parameter reuse, while occurring across different Transformer layers, whereas CoE reuses experts within the same layer.

\section{Experimental Setup}
\label{sec:exp}

\vspace{-3pt}
\subsection{Dataset and Metrics}
\vspace{-3pt}
We conduct our experiments on the general open-domain SlimPajama~\cite{cerebras2023slimpajama} and the reasoning domain-specific MetaMathQA~\cite{metamathqa} datasets. 
We train models on them separately and jointly (1:1 proportion) to evaluate the effect of sequential expert communication across domains. To assess the generalization performance, we track validation loss during training to measure optimization efficiency and convergence speed, and evaluate zero-shot performance on four standard benchmarks which span reasoning and commonsense tasks, including ARC-E~\cite{arc}, HellaSwag~\cite{hellaswag} and PIQA~\cite{piqa}, with (normalized or soft) answer accuracy\footnote{https://github.com/EleutherAI/lm-evaluation-harness/issues/1396}. 
\vspace{-3pt}
\subsection{Training Settings}
\vspace{-3pt}
Training is performed using the AdamW optimizer with a learning rate of 3e-4, weight decay of 0.01, and betas set to (0.9, 0.95). We use a linear learning rate schedule with 10\% warmup. Models are trained using a batch size of 64 and a sequence length of 512. To stabilize training, we apply gradient clipping with a threshold of 1.0.

\subsection{Model Configuration}
\vspace{-3pt}
Our model is built upon the \textbf{DeepSeek-V2-Lite}~\cite{deepseekv2} architecture, following the same design while scaling down to a smaller configuration with 544 million parameters (excluding embeddings) to facilitate faster convergence and enable efficient experimental validation. The transformer backbone comprises 4 layers, each with hidden size 1024 and 8 attention heads. All layers are configured as MoE layers with a total of 63 routed experts and 1 shared expert. For each token, the router selects 8 routed experts and the shared expert, and each expert has an intermediate size of 704. In our CoE setup, we apply 2 iterations of expert processing with inner residual connections and enable independent gating per iteration. This configuration ensures a fair comparison between traditional MoE and CoE under similar parameter budgets and architectural depth.
\vspace{-3pt}
\subsection{System and Infrastructure}
\vspace{-3pt}
All experiments are implemented within PyTorch, and we also borrow the Fully-Sharded Data Parallel (FSDP) Trainers from the \textbf{veRL} framework\footnote{\url{https://github.com/volcengine/verl}}~\citep{sheng2024hybridflow}, extending them to support multi-round expert execution and fine-grained token-level logging. The full detailed experimental code will be open-sourced soon. We conduct training on NVIDIA H100 GPUs, and each run completes within one GPU hour, enabling reproducibility without the need for large-scale compute infrastructure.

\section{Experimental Results and Analysis}
\vspace{-5pt}

In this section, we discuss several research questions that guide our investigation into how CoE may enable more effective language modeling:
\begin{enumerate}[noitemsep, topsep=0pt, leftmargin=*]
    \item Section~\ref{sec:sequential}: under the same compute budget, can the \textit{communicative token processing} mechanism enhance language modeling compared to parallel processing?
    \item Section~\ref{sec:scaling}: when scaling up computation, can expert iteration could serve as a better scaling factor compared to existing factors, such as model depth, width, expert choice count?
    \item Section~\ref{sec:keyfactor}: what design choices could make sequential expert processing effective?
    \item Section~\ref{sec:activation}: does the expert communication mechanism enable step-wise expert specialization?
    \item Section~\ref{sec:theory}: does chain-of-expert processing exhibit a theoretical advantage in combinatorics and representational flexibility that explains its improved efficiency?
\end{enumerate}
We explore these questions in the following subsections below. We also investigate the effect of shared experts in the Supplementary.

\begin{table}[t]
\centering
\footnotesize
\caption{\textbf{General benchmark performance comparison between CoE and MoE}. Under fixed expert compute across training sources, CoE ($K{=}4$, $C{=}2$) generally achieves better performance than MoE ($K{=}8$, $C{=}1$) on most settings, especially when trained and evaluated on mathematical reasoning datasets.}
\vspace{3pt}
\setlength{\tabcolsep}{4pt}
\vspace{0.1in}
\begin{tabular}{llcccccc}
\toprule
\textbf{Training Dataset} & \textbf{Model} & \multicolumn{2}{c}{\textbf{ARC-E}} & \multicolumn{2}{c}{\textbf{HellaSwag}} & \multicolumn{2}{c}{\textbf{PIQA}} \\
& & Acc & Norm & Acc & Norm & Acc & Norm \\
\midrule
\multirow{2}{*}{\textbf{SlimPajama}} 
& MoE $K{=}8$, $C{=}1$  & 27.2\% & 26.4\%  & 25.8\% & 25.1\% & 52.9\% & 51.0\% \\
& CoE  $K{=}4$, $C{=}2$ & \textbf{28.1\%} & \textbf{26.8\%} & \textbf{26.0\%} & \textbf{25.1\%} & \textbf{53.1\%} & \textbf{51.1\%} \\
\midrule
\multirow{2}{*}{\textbf{MetaMathQA}} 
& MoE $K{=}8$, $C{=}1$  & 26.4\% & 25.8\% & \textbf{26.1\%} & 26.0\% & \textbf{54.6\%} & \textbf{52.3\%} \\
& CoE $K{=}4$, $C{=}2$  & \textbf{26.5\%} & \textbf{26.0\%} & 25.9\% & \textbf{26.3\%} & 54.5\% & 51.4\% \\
\midrule
\multirow{2}{*}{\textbf{Combined Training}} 
& MoE $K{=}8$, $C{=}1$  & 26.4\% & 28.0\% & 26.5\% & 26.4\% & \textbf{56.2\%} & \textbf{54.2\%} \\
& CoE  $K{=}4$, $C{=}2$ & \textbf{27.7\%} & \textbf{28.2\%} & \textbf{26.5\%} & \textbf{26.8\%} & 55.2\% & 53.3\% \\
\bottomrule
\end{tabular}
\vspace{3pt}
\label{tab:coevsmoe_summary}
\end{table}

\begin{figure}[t]
    \begin{subfigure}[b]{0.48\linewidth}
        \centering
        \includegraphics[width=\linewidth]{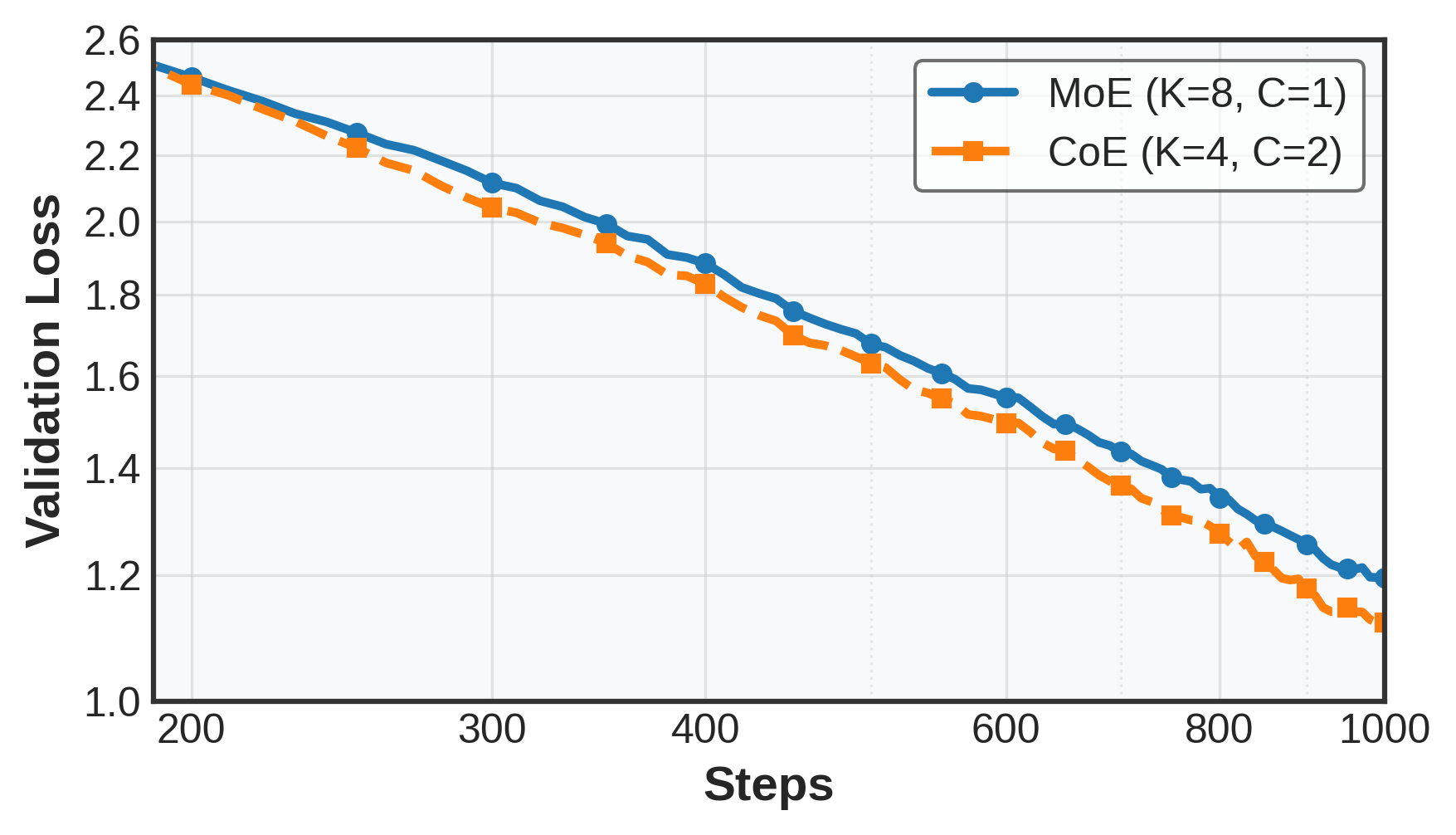}
        \caption{Validation loss with same expert budget.}
        \label{fig:loss}
    \end{subfigure}
    \hfill
    \begin{subfigure}[b]{0.48\linewidth}
        \centering
        \includegraphics[width=\linewidth]{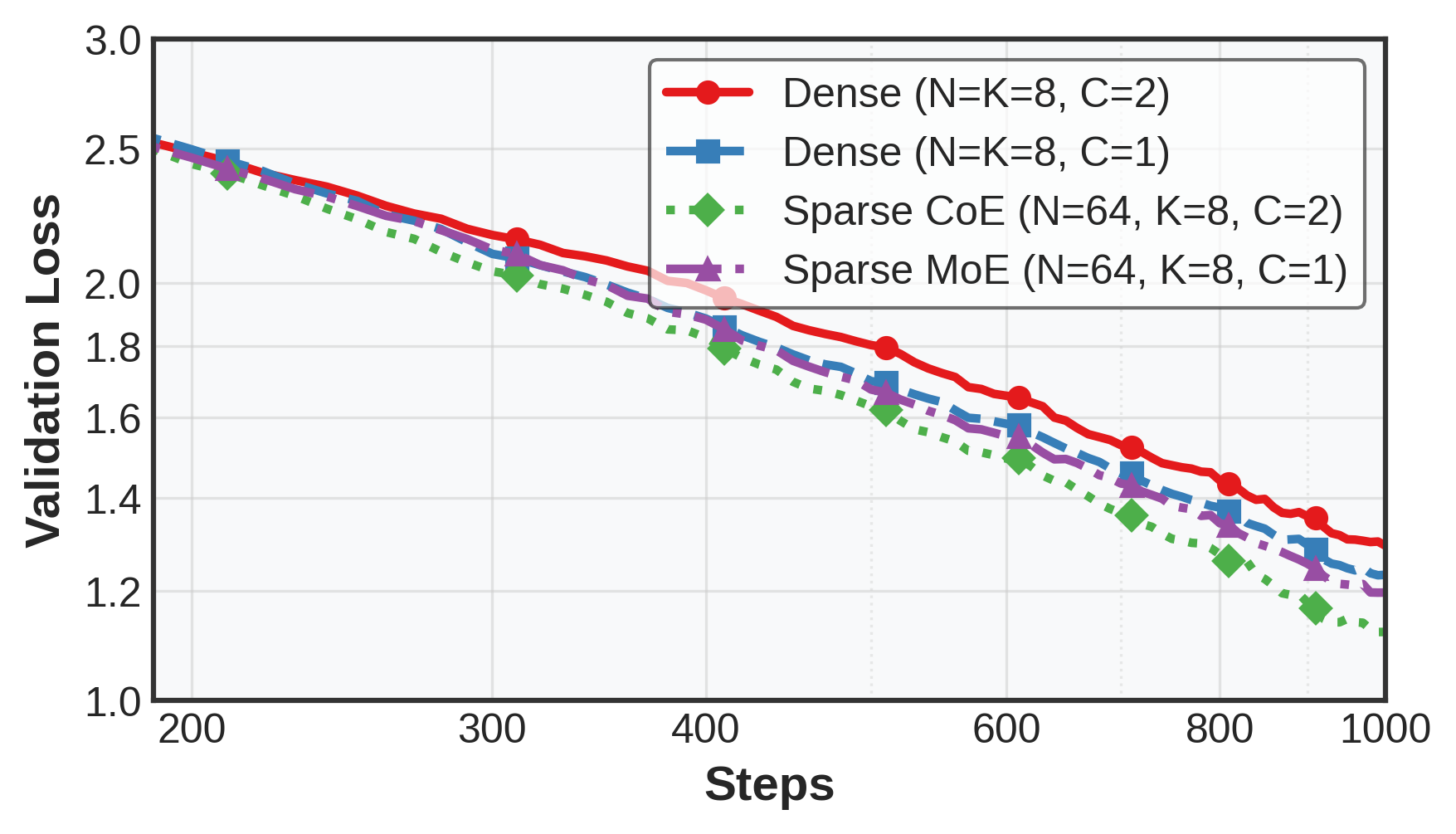}
        \caption{Iteration effect under sparse and dense setting.}
        \label{fig:sparsevsdense}
    \end{subfigure}
    \label{fig:coevsmoe}
    \caption{\textbf{CoE reduces validation loss more effectively than MoE under equal expert compute, specifically in sparse settings.} CoE ($K{=}4$, $C{=}2$) outperforms MoE ($K{=}8$, $C{=}1$) with the same per-token expert processing (left). The benefit is amplified in sparse routing, where communication fosters specialization, but diminishes in dense settings where all experts are active (right).}
    \vspace{-5pt}
\end{figure}

\vspace{-3pt}
\subsection{Communicative Processing can be Better than Parallel Processing in MoEs}
\label{sec:sequential}
\vspace{-3pt}
We begin by discussing whether communicative processing could lead to better language modeling by setting a fixed expert-processing count at 8 per layer, adapting the communication steps $C$ to explore whether models can benefit from expert communication. As shown in Figure~\ref{fig:loss}, CoE ($C=2, K=4$) consistently outperforms MoE ($K=8$) with equal expert processing per token, reducing validation loss from 1.20 to 1.12 while exhibiting faster convergence.
This advantage also generalizes beyond loss curves: as summarized in Table~\ref{tab:coevsmoe_summary}, CoE outperforms or achieves comparable performance as MoE on multiple downstream benchmarks across ARC-E,
HellaSwag, PIQA. Due to the small model size, the comparison has not yet shown a significant gap, but it has already demonstrated the potential of CoE under different settings.

Interestingly, we find that CoE’s advantage is primarily realized in \textbf{sparse} configurations. As shown in Figure~\ref{fig:sparsevsdense}, when the model selects only a subset of experts per step (sparse routing), increasing the number of communication steps $C$ leads to a noticeable improvement in validation loss. In contrast, under \textbf{dense} settings where all experts are always active (e.g., Total=$K$=$8$, $C$=$2$), iterative processing provides limited gain over one-shot routing. We hypothesize that this is because sparsity encourages expert specialization, allowing different iterations to focus on refining different token aspects. Without sparsity, the repeated processing simply deepens the computation path without introducing additional diversity, diminishing the benefit of communication.
We further investigate this phenomenon in Section~\ref{sec:activation}, where we show that different iteration lead to diverse expert sets.

\begin{figure}[t]
    \centering
    \includegraphics[width=\linewidth]{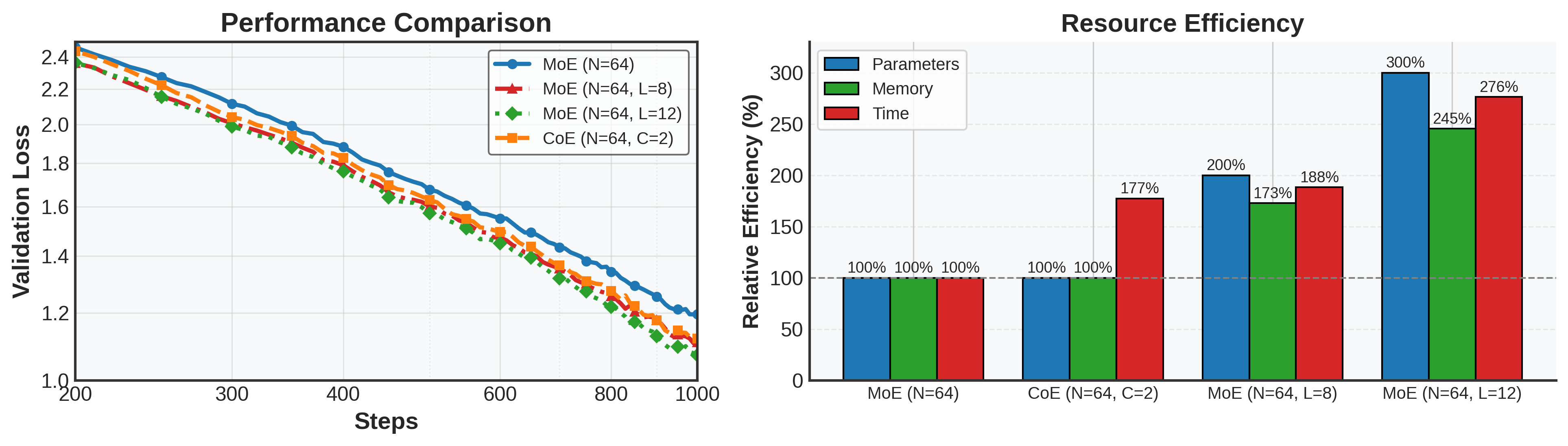}
    \caption{\textbf{Depth scaling comparison.} CoE ($C{=}2$) with 4 layers matches the performance of deeper MoE models ($L$=$8$ or $12$) with significantly lower memory and parameter usage.}
    \label{fig:depth}
\end{figure}

\begin{figure}[t]
    \centering
    \includegraphics[width=\linewidth]{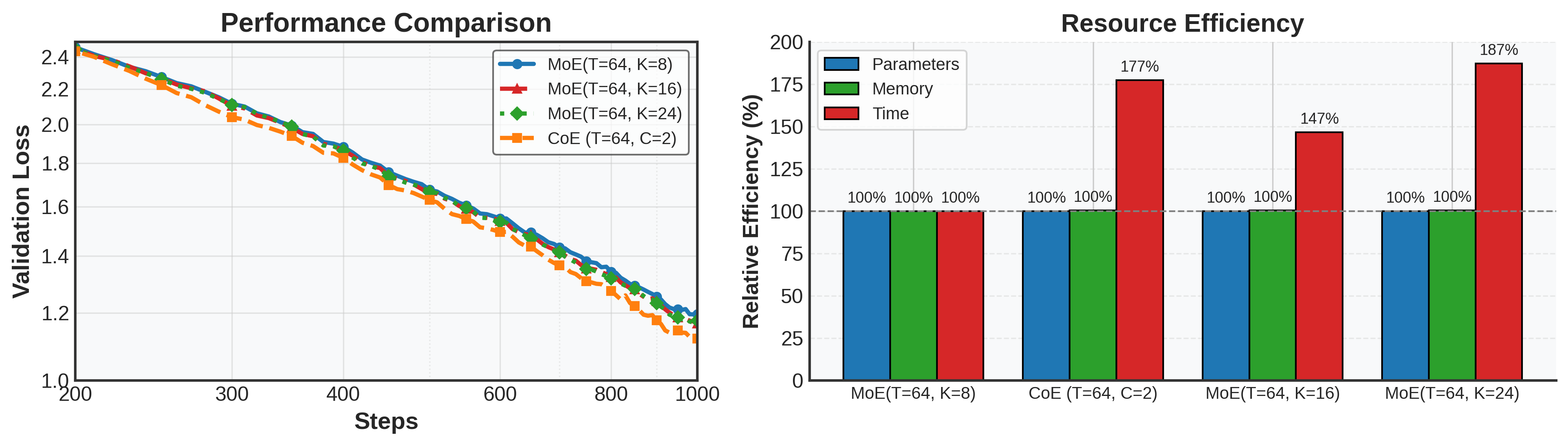}
    \caption{\textbf{Width scaling comparison.} CoE ($C{=}2$) outperforms MoE variants with increased expert selection ($K{=}16$ or $24$), while using similar resources.}
    \label{fig:width}
\end{figure}

\vspace{-3pt}
\subsection{Expert Communication Steps can Serve as a Better Scaling Factor}
\label{sec:scaling}
\vspace{-3pt}
To confirm the scaling effect of expert communication, we compare the communication step count ($C$) in CoE to conventional scaling dimensions in MoE architectures, such as network depth and expert selection width. Across three controlled experiments on the Math domain, we find that increasing communication steps in CoE can offer a more efficient scaling path, gaining comparable or better performance with lower memory and compute overhead.

\paragraph{(a) CoE can match deeper MoE with lower memory cost.}
To control for network depth, we fix the number of experts ($N{=}8$) and the number of experts selected per token ($K{=}8$), and scale the number of transformer layers from 4 to 8 and 12. We compare these models to a 4-layer CoE model with $C{=}2$ expert communication steps. As shown in Figure~\ref{fig:depth}, CoE achieves similar performance to MoE ($L$=$12$) while reducing memory usage by 42\%. Unlike deeper MoE variants, CoE maintains constant parameter count and layer depth, reducing memory overhead with similar training time.

\paragraph{(b) CoE can outperform wider MoE under equal selection budget.}
We next scale the width of expert selection by varying the chosen experts each layer $K$ from 8 to 16 and 24 in standard MoE, while keeping $T{=}64$ and $L{=}4$. In contrast, CoE retains $K{=}8$ but introduces communication ($C{=}2$). As shown in Figure~\ref{fig:width}, CoE achieves better convergence while consuming comparable memory and compute. This indicates that increasing $C$ could be a more effective way to expand expert processing than simply increasing $K$.

\begin{figure}[t]
    \centering
    \includegraphics[width=\linewidth]{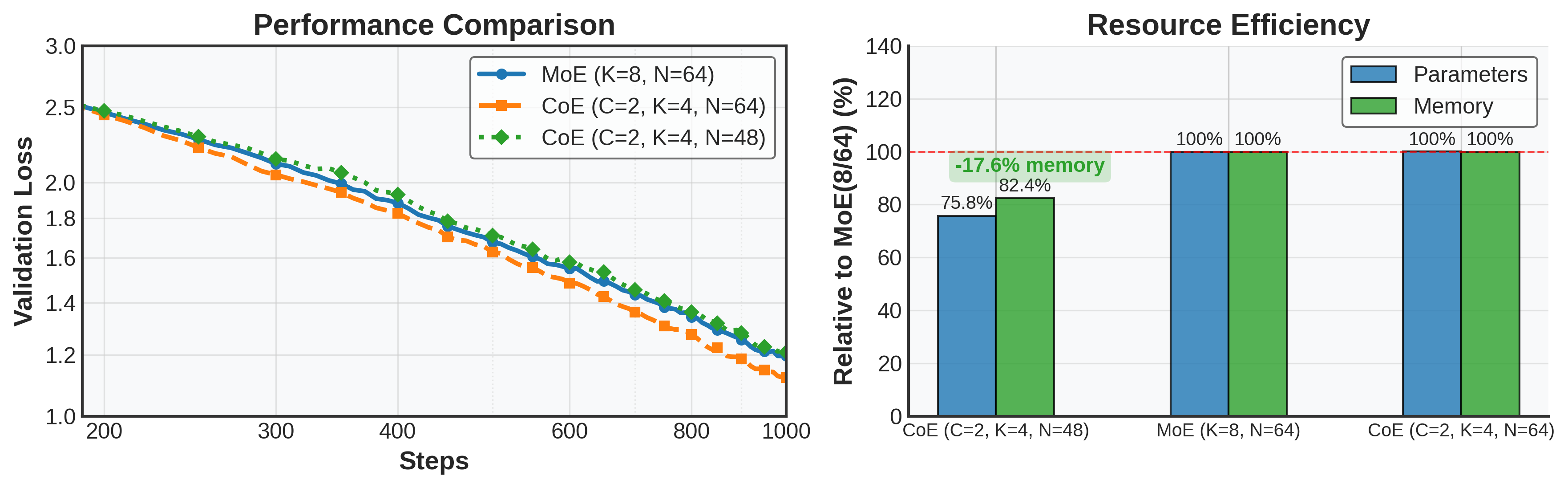}
    \caption{\textbf{Matched performance with fewer experts.} CoE ($N{=}48$) achieves similar performance to MoE ($N{=}64$) while reducing memory usage by \textbf{17.6\%}.}
    \label{fig:efficiency}
\end{figure}

\paragraph{(c) CoE delivers matched performance with fewer total experts.}
We further present that CoE can reduce the total number of experts while preserving performance. In Figure~\ref{fig:efficiency}, CoE with $C{=}2$, $K{=}4$, and $N{=}48$ achieves similar validation loss to MoE with $K{=}8$ and $N{=}64$, but reduces memory usage by \textbf{17.6\%}. This demonstrates that CoE can achieve compute-efficient generalization by improving expert reuse instead of relying on scale alone.

Together, these results suggest that expert communication steps ($C$) in CoE offer a more efficient and scalable way to expand model capacity. Unlike traditional depth or width scaling which often increases memory footprint and compute cost, scaling $C$ improves expert reuse and compositional depth without growing parameter count or memory usage.

\subsection{Key Components Enabling Effective Sequential Expert Processing}
\label{sec:keyfactor}

To understand what makes sequential expert routing effective in CoE, we conduct an ablation study focusing on two key architectural choices: iteration-specific gating and inner residual connections. Both components are designed to improve expert compositionality and training stability. We compare ablated variants against standard CoE and a strong MoE baseline.

\paragraph{Iteration-specific gating enables per-step specialization.}
A central hypothesis of CoE is that each communication step should dynamically route tokens to different experts based on their evolving representation. To test this, we construct a variant that removes iteration-specific gating and instead reuses the same expert selection across all steps. Formally, the update rule becomes:
\begin{equation}
\begin{aligned}
x^{(0)} &= x, \\
x^{(t)} &= \sum_{i=1}^{M} \hat{\text{E}}_i(x^{(t-1)}) + \sum_{i=1}^{N} g_i \cdot \text{E}_i(x^{(t-1)}) + \mathbb{I}_r \cdot x^{(t-1)}, \quad t = 1, \ldots, C, \\
y &= x^{(C)},
\end{aligned}
\end{equation}
where the gating weights $g_i$ are fixed across iterations. This design removes the model’s ability to condition expert routing on intermediate computation states.

\begin{figure}[t]
    \centering
    \includegraphics[width=0.6\linewidth]{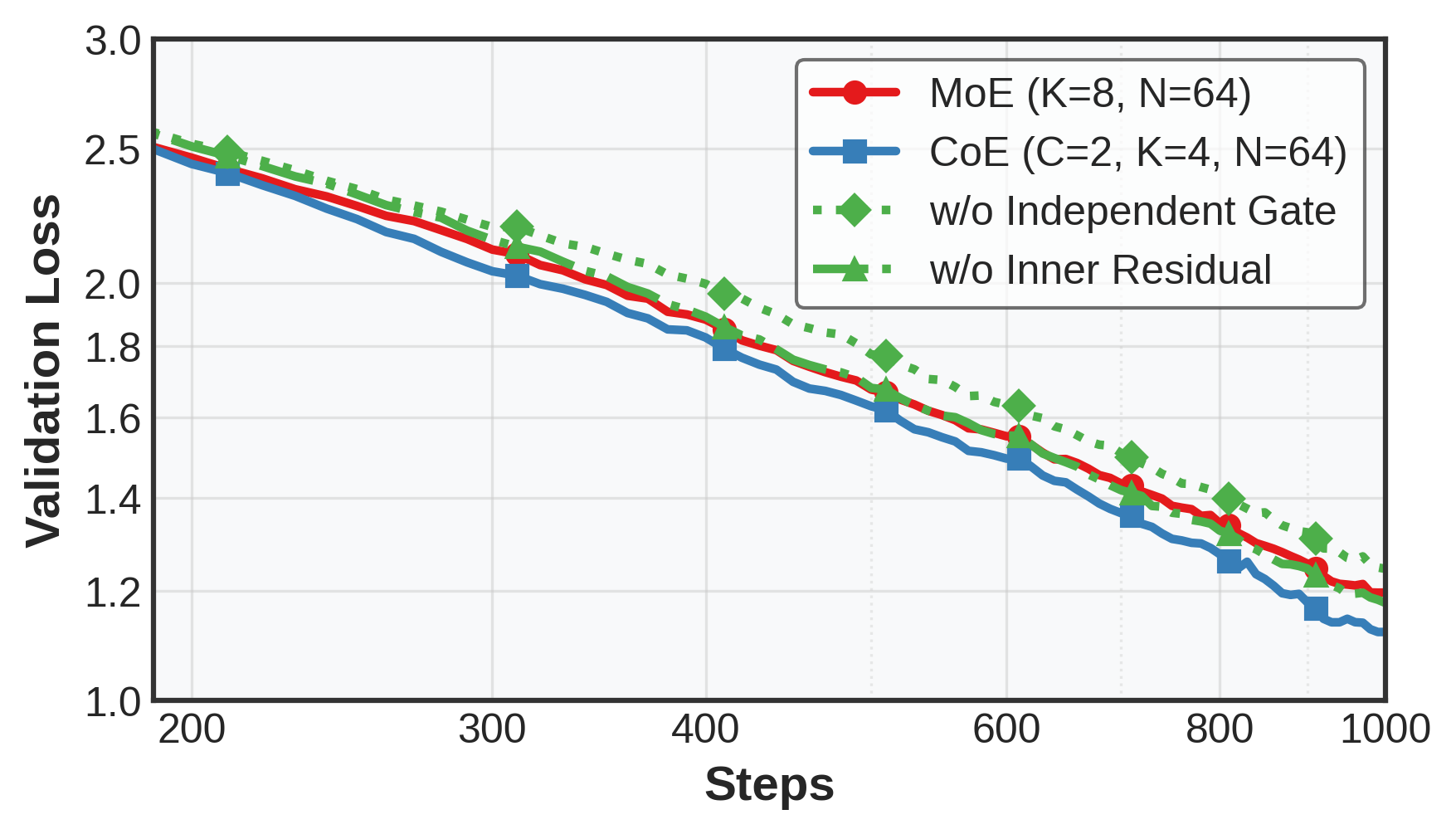}
    \caption{\textbf{Ablation study.} Removing either iteration-specific gating or inner residuals significantly reduces performance. Both components are critical to the effectiveness of sequential expert routing in CoE.}
    \label{fig:abl}
\end{figure}

As shown in Figure~\ref{fig:abl}, this shared-gating variant significantly underperforms: its validation loss quickly plateaus around 1.5, worse than both standard CoE and the MoE baseline. These results highlight the importance of step-specific routing in CoE—without it, the model fails to leverage the compositional benefits of iterative reasoning.

\paragraph{Inner residuals stabilize multi-step refinement.}
We also investigate how residual connections should be applied across communication steps. In standard CoE, residuals are applied at each iteration—referred to as \textit{inner residuals}—to support progressive token-wise refinement. An alternative design applies a single \textit{outer residual} after all iterations, skipping intermediate feedback. Formally, this variant modifies the update rule as:
\begin{equation}
\begin{aligned}
x^{(0)} &= x, \\
x^{(t)} &= \sum_{i=1}^{M} \hat{\text{E}}_i(x^{(t-1)}) + \sum_{i=1}^{N} g_{t,i} \cdot \text{E}_i(x^{(t-1)}), \quad t = 1, \ldots, C, \\
y &= x^{(C)} + x,
\end{aligned}
\end{equation}
where only the final output receives residual feedback from the original input. As shown in Figure~\ref{fig:abl}, this outer-residual-only design leads to slower convergence and higher final validation loss when trained on mathematical data. These results suggest that inner residuals play a key role in stabilizing multi-step reasoning, allowing more effective credit assignment and optimization along the iterative expert path.

Together, these results confirm that CoE’s performance gains stem not only from multi-step routing, but also from the architectural mechanisms that support specialization via independent gating and stable refinement via inner residual connections.

\subsection{Co-activation Patterns Reveal Step-wise Expert Specialization}
\label{sec:activation}
\begin{figure}
    \centering
    \includegraphics[width=1\linewidth]{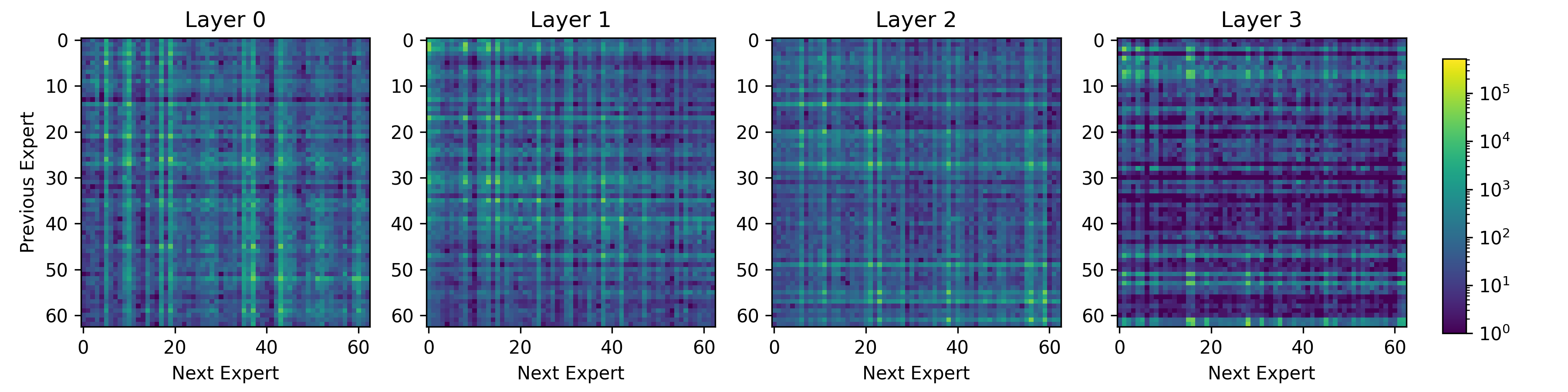}
    \caption{\textbf{Expert co-activation between communication steps.} For each selected layer, we plot a matrix counting how often each expert pair $(e_0, e_1)$ is activated across the two communication steps in CoE. The non-uniform, asymmetric patterns indicate that routing decisions differ meaningfully between steps, supporting the emergence of step-wise expert specialization.}
    \label{fig:routing_matrix_all_layers}
\end{figure}

To further understand how communication steps affect routing behavior, we visualize the co-activation matrix between the two iterations of CoE. For each token routed to experts in iteration 0 and 1, we accumulate expert pair frequencies, resulting in a $K \times K$ co-activation matrix per layer. Figure~\ref{fig:routing_matrix_all_layers} shows the matrices across four layers trained and evaluated on MetaMathQA, and we present more settings for this expert co-activation in the Supplementary.

We observe that expert pairs are not uniformly distributed: each layer exhibits a diverse set of expert combinations across steps, suggesting that different iterations route to meaningfully different experts. This supports our hypothesis that CoE leverages iteration-specific gating to perform progressive, compositional refinement. Moreover, the sparsity and asymmetry of the co-activation matrices imply that the model is not simply repeating expert usage, but adapting routing based on intermediate representations.

Together with the results in Figure~\ref{fig:abl} and Table~\ref{tab:coevsmoe_summary}, this analysis highlights that CoE achieves more than deeper computation—it enables structured multi-step specialization that standard MoE cannot capture.

\subsection{Theoretical Analysis: Combinatorial Flexibility and Effective Depth}
\label{sec:theory}

We hypothesize the performance advantages of CoE could stem from two properties: combinatorial flexibility in expert selection and implicit depth expansion through expert communication.
Conventional MoEs that select $2k$ experts in a single step, while CoE performs two separate top-$k$ routing operations across iterations. This change increases the number of possible expert combinations from $C(n, 2k)$ to $C(n, k)^2$. For example, with $n = 64$ and $k = 4$, CoE leads to $823\times$ more expert pairings than one-shot routing. It allows the model to encode a significantly more diverse set of expert interactions, which may improve its ability to route tokens in a more diverse manner.

In addition, CoE introduces iterative processing that deepens token representations over time. Since expert outputs from the first iteration influence routing in the second, the model applies distinct transformations across steps. A token may be refined by different experts or revisited by the same expert, enabling multi-pass representation refinement. This increases the model's effective depth without increasing parameter count or layer count. Recent analyses~\citep{feng2023revealing,merrill2025littledepthgoeslong,Geiping2025LatentReasoning} suggest that deeper internal computation pathways correlate with improved reasoning, particularly in math and logical inference. By enabling step-wise expert composition, CoE supports this depth-like refinement within a sparse, modular architecture.

\section{Related Works}

\textbf{Mixture of Experts (MoE).}~~
MoE improves scaling efficiency by activating sparse subnetworks~\citep{jacobs1991adaptive,shazeer2017outrageously}. Recent MoE-based LLMs such as Mixtral~\citep{jiang2024mixtral}, DeepSeek~\citep{deepseekmoe,deepseekv2,liu2024deepseekv3,esft}, Phi-3.5~\citep{abdin2024phi3}, and Qwen3~\citep{qwen2025qwen3} adopt top-$k$ expert routing to reduce active parameter count during inference. These models maintain competitive performance with significantly lower memory cost. The architecture has also been adopted in Llama 4~\citep{meta2025llama4} and explored in OLMoE~\citep{muennighoff2024olmoe} and Skywork~\citep{wei2024skywork}.

\textbf{Iterative Architectures and Effective Depth.}~~
Instead of increasing width, iterative (recurrent) models reuse layers iteratively to expand effective depth~\cite{bansal2022endtoend, bear2024rethinkingdeepthinkingstable, mcleish2024transformers}. Universal Transformers~\citep{Dehghani2019Universal} and ACT~\citep{Graves2016ACT} use input-adaptive computation steps; ALBERT~\citep{Lan2020ALBERT} and DEQ~\citep{Bai2019DEQ} share or implicitly unroll layers. Recent work explores inference-time depth via latent recurrence~\citep{Geiping2025LatentReasoning}, offering computation-quality tradeoffs beyond prompt-based reasoning~\citep{Wei2022CoT}. This setup improves reasoning controllability, while remaining robust under compression~\citep{Zhang2025CompressionReasoning}.

\textbf{Expert Reuse and Interaction.}~~
Recent work enhances expert capacity via iterative reuse. SparseUT~\citep{tan2023sparse} and MoEUT~\citep{csordas2024moeut} combine MoE with depth-shared layers. RMoE~\citep{qiu2024layerwise} introduces a recurrent routing state. OLMoE~\citep{muennighoff2024olmoe} mixes global and token-wise expert selection. While most designs apply inter-layer reuse, intra-layer expert communication remains underexplored.

\section{Conclusion}

Chain-of-Experts (CoE) improves sparse model performance by introducing iterative expert processing within each transformer layer. Unlike traditional Mixture-of-Experts (MoE) architectures that treat expert computation as independent, CoE applies sequential routing with separate gating at each step, allowing experts to refine and exchange information across iterations. Empirical results show that CoE achieves lower validation loss, more efficient use of memory, and better compute scaling under matched parameter and runtime budgets. These findings support the view that sparse model quality depends not only on the number of experts used, but also on how information flows between them, pointing to a potentially better sparse foundation model architecture where modular reasoning depends on expert communication rather than solely expert capacity.

\section{Limitations and Future Work}
\label{sec:conclusions}
While Chain-of-Experts (CoE) demonstrates strong performance and efficiency, several practical and conceptual limitations remain. CoE introduces moderate time overhead in practice despite theoretical FLOP parity to MoE due to sequential processing. Selecting fewer experts per iteration reduces the degree of matrix multiplication parallelism, which can slow training on current hardware unless optimized at a low level. CoE also requires training from scratch and is not compatible with existing pretrained MoE checkpoints, limiting its use in transfer learning workflows. In addition, our implementation has only been tested on single-device setups. It remains unclear how CoE behaves under multi-node training or when used with newer training formats such as FP8.

Future work will focus on scaling model size, batch size, and training steps to examine whether CoE's advantages persist under standard scaling law conditions. Although our current experiments primarily use math datasets due to their challenging reasoning nature, broader domain evaluation is necessary. We will also study CoE's performance on downstream tasks, including language understanding and code generation benchmarks. Another direction is to increase the number of expert processing iterations per layer. Our current experiments use only two; the effect of deeper iterative depth remains unexplored. Finally, how to combine CoE with architectural variants that share experts across layers such as MoEUT remains an open-problem. These designs replace within-layer routing with global expert pools shared across the network, potentially enabling greater parameter efficiency and richer specialization patterns.

\bibliography{custom} 
\bibliographystyle{unsrtnat}

\clearpage

\appendix
\section{Extended Related Works}

\paragraph{Mixture-of-Experts (MoE).}
MoE architectures extend neural capacity through conditional computation~\citep{jacobs1991adaptive}, activating only a small subset of experts per input~\citep{shazeer2017outrageously}. Modern MoE-based LLMs like Mixtral~\citep{jiang2024mixtral} adopt top-2 routing across 8 feed-forward experts, reducing memory footprint during inference. This pattern enables models to scale parameter counts without increasing runtime cost. Several open models including DeepSeek~\citep{deepseekmoe,deepseekv2,liu2024deepseekv3,esft}, Phi-3.5~\citep{abdin2024phi3}, Qwen3~\citep{qwen2025qwen3}, and Llama 4~\citep{meta2025llama4} demonstrate the practicality of MoE in production-grade systems. Complementary efforts such as OLMoE~\citep{muennighoff2024olmoe} and Skywork~\citep{wei2024skywork} also explore hybrid routing schemes and inference-time budget control.

\paragraph{Iterative Computation and Effective Depth.}
Iterative processing architectures offer an alternative scaling axis by reusing modules across time~\citep{bansal2022endtoend, bear2024rethinkingdeepthinkingstable, mcleish2024transformers, ELMAN1990179}. Early work such as~\citet{Schmidhuber1992FastWeight} and~\citet{Hochreiter1997LSTM} emphasized iteration (recurrence) as a means for structured memory and temporal abstraction. Modern models like Universal Transformers~\citep{Dehghani2019Universal} and Adaptive Computation Time (ACT)~\citep{Graves2016ACT} dynamically control depth through halting units, while ALBERT~\citep{Lan2020ALBERT} and~\citet{Takase2023ParameterSharing} achieves parameter efficiency via cross-layer sharing. Implicit infinite-depth computation is enabled by deep equilibrium models (DEQ)~\citep{Bai2019DEQ}, and recent dynamic-depth strategies include early-exit~\citep{Elbayad2020DepthAdaptive}, LayerDrop~\citep{Fan2020LayerDrop}, and residual depth scaling~\citep{kim2024solar107bscalinglarge}.
More recently, latent-space recurrence has been explored as a lightweight mechanism for test-time refinement:~\citet{Geiping2025LatentReasoning} show that recursive hidden-state iteration improves math and logical reasoning. These methods internalize multi-step computation, in contrast to chain-of-thought prompting~\citep{Wei2022CoT}. They also tend to be more compression-resilient~\citep{Zhang2025CompressionReasoning}, consistent with findings on the robustness of recurrent representations~\citep{Csordas2021DevilDetail, schmidhuber2012selfdelimitingneuralnetworks}.

\paragraph{Expert Reuse and Interaction.}
In addition to sparse activation, recent research focuses on improving expert utility through reuse and structural regularization. SparseUT~\citep{tan2023sparse} and MoEUT~\citep{csordas2024moeut} integrate MoE into Universal Transformers, enabling both depth-sharing and expert reuse. MoEUT further unifies attention and FFN expert selection and adds normalization to stabilize cross-layer reuse. Layerwise recurrent MoE (RMoE)~\citep{qiu2024layerwise} introduces a recurrent router state, enhancing context-aware routing decisions. DeepSeekMoE~\cite{deepseekmoe} and OLMoE~\citep{muennighoff2024olmoe} separates experts into global and token-routed groups to promote common feature sharing. However, these systems mostly adopt inter-layer sharing. Intra-layer expert communication remains relatively unexplored, opening opportunities for expert collaboration across time steps and interaction loops, an idea central to our Chain-of-Experts formulation.

\section{Impact of Shared Experts}

Our framework allows for flexible inclusion of shared experts across both CoE and MoE variants. To assess their utility, we conduct a controlled comparison using four configurations: CoE with and without a shared expert, and MoE with and without a shared expert. All these models are trained on the MetaMathQA dataset.

As shown in Table~\ref{tab:shared_expert_analysis}, introducing a shared expert leads to higher PIQA normalization scores. These gains are more pronounced in the smaller $K{=}4$, $C{=}2$ configuration, suggesting that shared experts help compensate for limited expert diversity by providing a stable backbone for generalization. On ARC-E and HellaSwag, however, performance remains largely similar, and in some cases models without shared experts slightly outperform their counterparts. It indicates that while shared experts can enhance performance on certain tasks, their benefit is not universal and may depend on task characteristics or expert routing depth.

\begin{table}[t]
\centering
\caption{\textbf{Impact of shared expert on CoE performance across benchmarks.} Using a shared expert yields marginal gains on PIQA for both MoE and CoE, while overall performance remains comparable across settings with and without shared experts.}
\vspace{3pt}
\setlength{\tabcolsep}{4pt}
\begin{tabular}{lcccccc}
\toprule
\textbf{Setting} & \multicolumn{2}{c}{\textbf{ARC-E}} & \multicolumn{2}{c}{\textbf{HellaSwag}} & \multicolumn{2}{c}{\textbf{PIQA}} \\
& Acc & Norm & Acc & Norm & Acc & Norm \\
\midrule
\textbf{MetaMathQA (1 shared)} & & & & & & \\
\quad $K{=}4$, $C{=}2$ & 26.5\% & \textbf{26.0}\% & 25.9\% & \textbf{26.3\%} & \textbf{54.5\%} & 51.4\% \\
\quad $K{=}8$, $C{=}1$ & 26.4\% & 25.8\% & 26.1\% & 26.0\% & \textbf{54.6\%} & \textbf{52.3\%} \\
\midrule
\textbf{MetaMathQA (0 shared)} & & & & & & \\
\quad $K{=}4$, $C{=}2$ & \textbf{27.1\%} & 25.6\% & \textbf{26.5\%} & 26.0\% & 52.2\% & 51.6\% \\
\quad $K{=}8$, $C{=}1$ & 26.7\% & 25.0\% & 26.2\% & \textbf{26.5\%} & 54.0\% & 52.2\% \\
\bottomrule
\end{tabular}
\label{tab:shared_expert_analysis}
\end{table}

\section{Extended Training and Analysis}

\subsection{Extending Training Steps} 
\begin{figure}[t]
    \centering
    \includegraphics[width=1\linewidth]{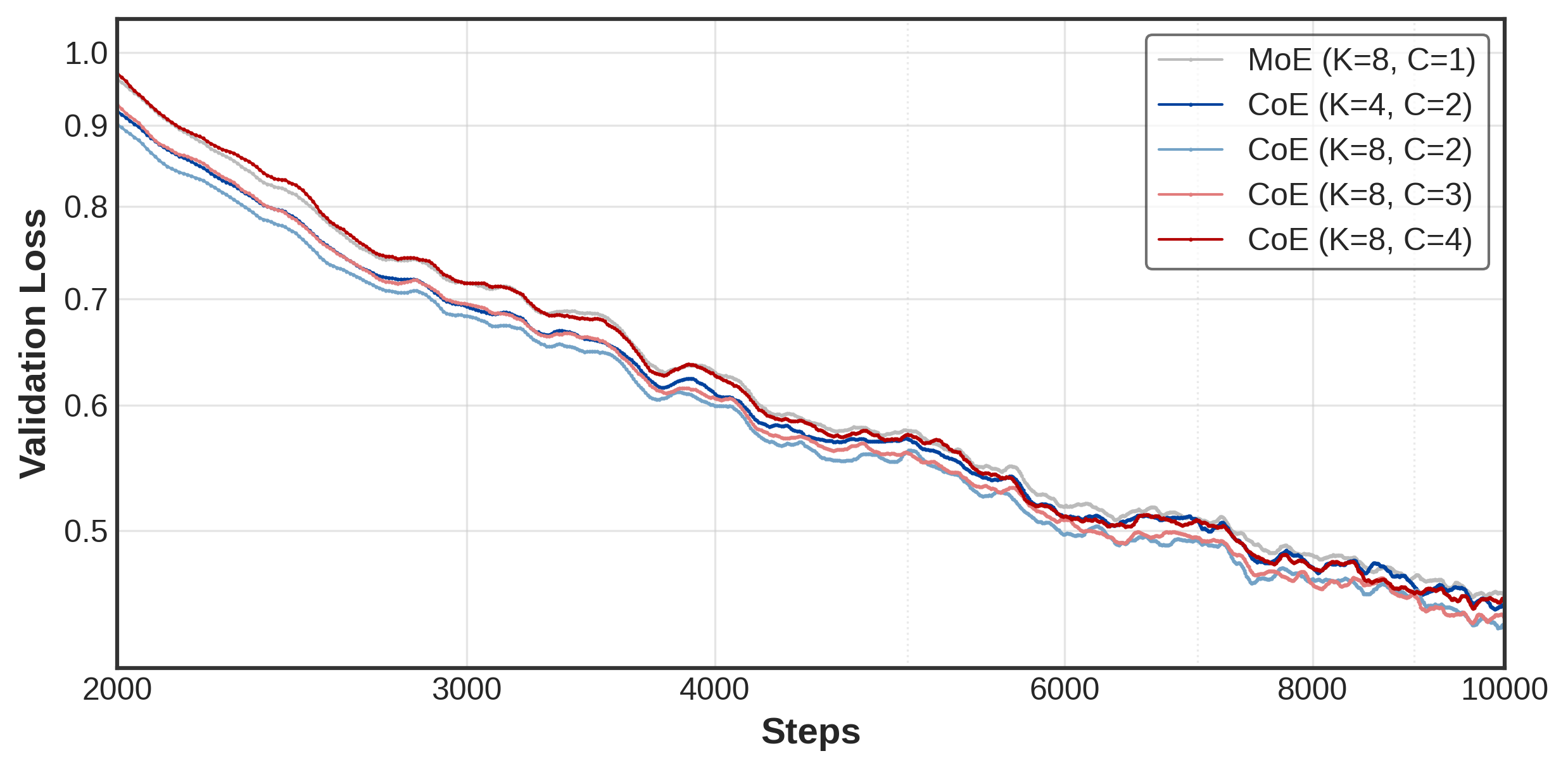}
    \caption{Validation loss on MetaMathQA across 10,000 steps.}
    \label{fig:plot_math}
\end{figure}

\begin{figure}[t]
    \centering
    \includegraphics[width=1\linewidth]{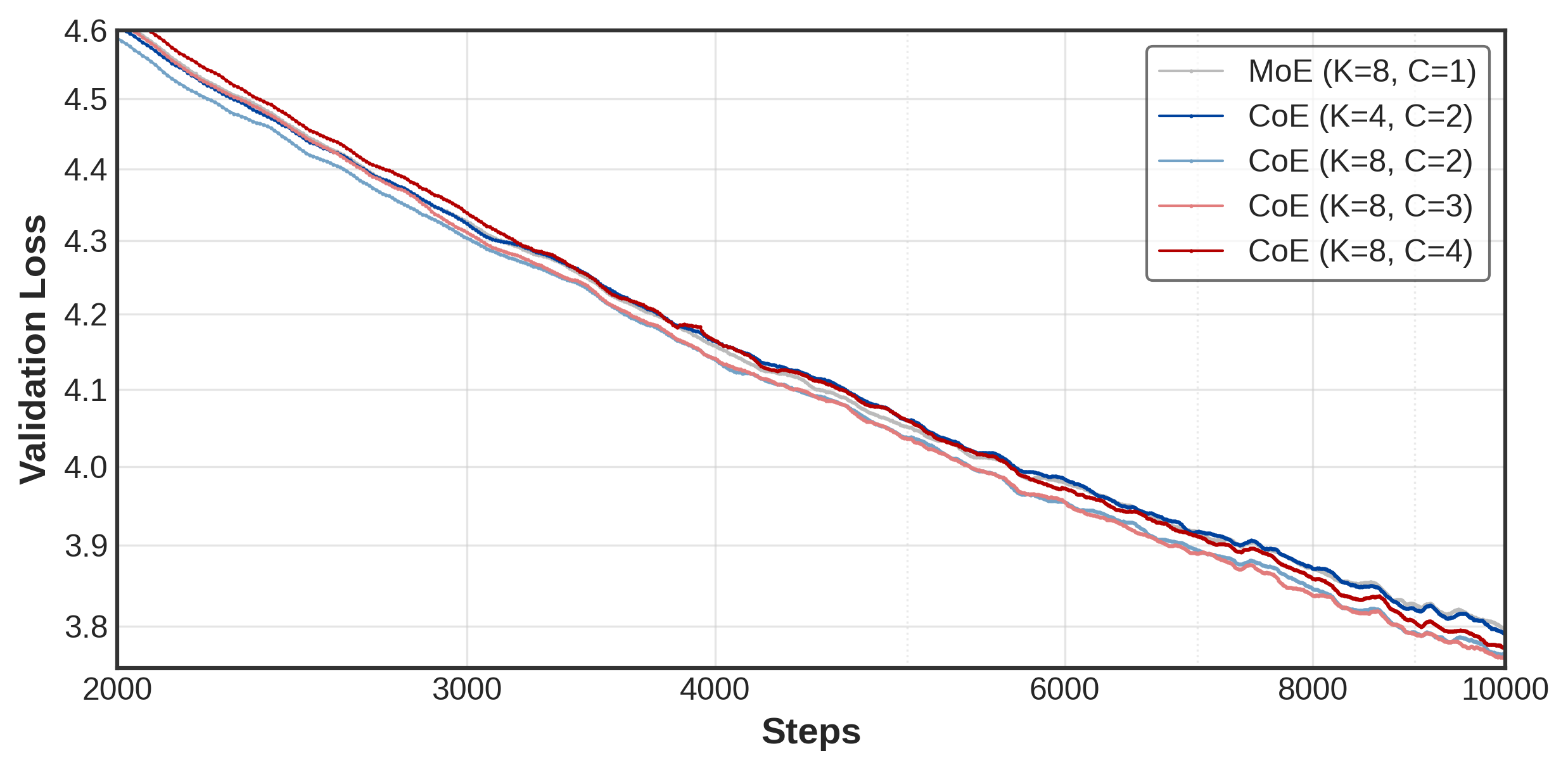}
    \caption{Validation loss on general-domain SlimPajama across 10,000 steps.}
    \label{fig:plot_slim}
\end{figure}

We extend training to 10{,}000 steps, 10\% warmup to examine whether earlier conclusions hold under increased compute. Note that compute on MoE layers is approximately proportional to \(C \times K\), representing the total number of expert invocations per token per layer. As shown in Figures~\ref{fig:plot_math} and~\ref{fig:plot_slim}, all models eventually converge, but the scaling behavior differs.

In both the MetaMathQA (Figure~\ref{fig:plot_math}) and the general-domain SlimPajama settings (Figure~\ref{fig:plot_slim}), CoE with \(K{=}8, C{=}2\) consistently achieves a lower final validation loss than MoE with \(K{=}8, C{=}1\). It indicates that increasing \(C\) in CoE offers a notable scaling factor. Additionally, CoE with \(K{=}4, C{=}2\) converges faster than MoE early in training, though its final loss is similar. It suggests that CoE benefits more from compute scaling, especially in the high-capacity regime, and that increasing \(C\) is an effective lever for improving convergence without increasing total parameters or GPU memory requirements.

\subsection{Extending Communication Steps}
To further explore the scaling behavior of CoE, we extend the communication steps $C$ beyond the previously studied $C=2$. Figures~\ref{fig:plot_math} and~\ref{fig:plot_slim} report validation loss curves under fixed $K=8$ and increasing $C \in \{2, 3, 4\}$. While increasing $C$ from 1 to 2 yields a clear improvement, further increasing $C$ shows diminishing or unstable gains, especially on the MetaMathQA benchmark.

Although CoE $(K=8, C=3/4)$ extends computation within a fixed parameter space, the performance only matches or slightly outperforms the baseline in early training, and the final convergence is less robust. It suggests that \textbf{naive scaling of $C$ may lead to inefficient communication or unstable optimization dynamics}, and may require more training compute to converge. Efficiently extending the communication horizon remains an open challenge, and we leave the development of more principled techniques for more effective expert communication to future work.

\subsection{Extended Analysis on Residual Connections}

In addition to the residual strategy proposed in Section~\ref{sec:keyfactor}, we explore an alternative residual design inspired by~\cite{Geiping2025LatentReasoning}, where the initial representation is added as a residual at every iteration. Formally, the update rule is defined as:
\begin{equation}
\begin{aligned}
x^{(0)} &= x, \\
x^{(t)} &= \sum_{i=1}^{M} \hat{\text{E}}_i(x^{(t-1)}) + \sum_{i=1}^{N} g_{t,i} \cdot \text{E}_i(x^{(t-1)}) + x^{(0)}, \quad t = 1, \ldots, C, \\
y &= x^{(C)}.
\end{aligned}
\end{equation}

We compare this variant (denoted as \textit{Init}) with two other designs: adding residuals from the previous iteration (\textit{Inner}) and from a separate outer loop representation (\textit{Outer}). All methods are trained for 1000 steps on the MetaMathQA dataset. The loss comparison is presented in Table~\ref{tab:residual-loss}.

\begin{table}[ht]
    \centering
    \caption{Residual loss under different residual connection strategies. }
    \vspace{3pt}
    \begin{tabular}{lccc}
        \toprule
        & Inner Residual & Outer Residual & Init Residual \\
        \midrule
        Loss & 1.12 & 1.21 & 1.18 \\
        \bottomrule
    \end{tabular}
        \label{tab:residual-loss}
\end{table}

These results suggest that residual connections must be carefully designed. While the \textit{Init} method provides some improvement over the \textit{Outer} strategy, it still underperforms compared to \textit{Inner} residual we used, which we adopt as the default in all main experiments.

\section{Analysis of Expert Co-Activation Visualizations}

\begin{figure}[H]
    \centering
    \includegraphics[width=\linewidth]{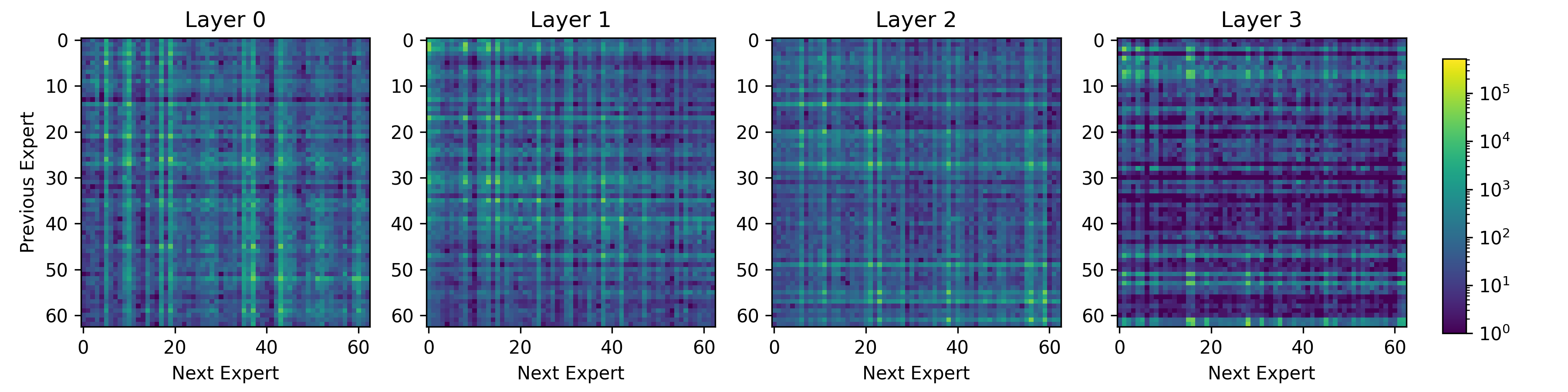}
    \caption{Layer-wise routing pattern on MetamathQA under the 4-experts-per-iteration, 2-iteration setup (2k training steps).}
    \label{fig:routing_4tpk2itr_metamathqa_2k}
\end{figure}

\begin{figure}[H]
    \centering
    \includegraphics[width=\linewidth]{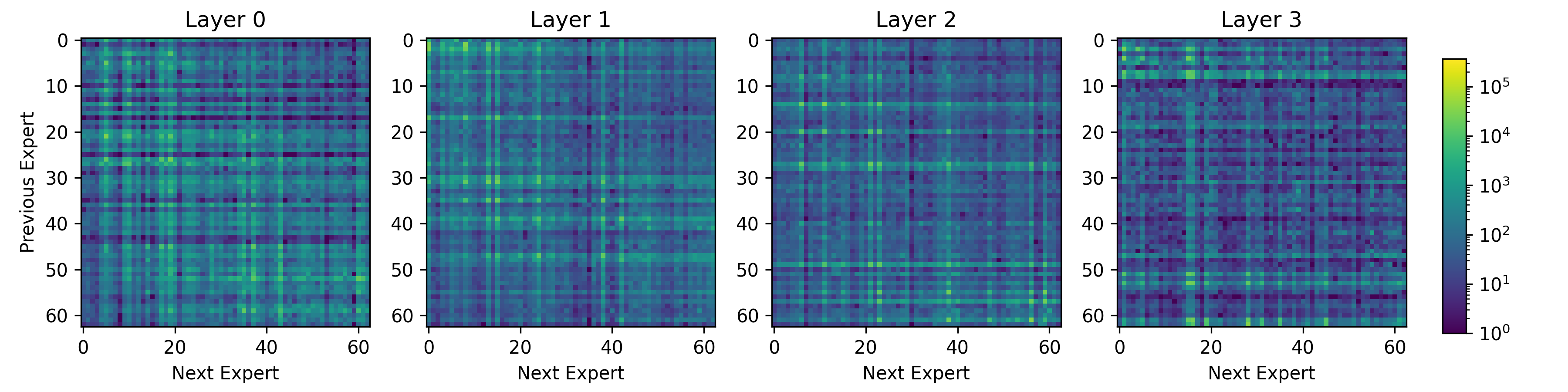}
    \caption{Layer-wise routing pattern on MetamathQA under the 4-experts-per-iteration, 2-iteration setup (10k training steps).}
    \label{fig:routing_4tpk2itr_metamathqa_10k}
\end{figure}

\begin{figure}[H]
    \centering
    \includegraphics[width=\linewidth]{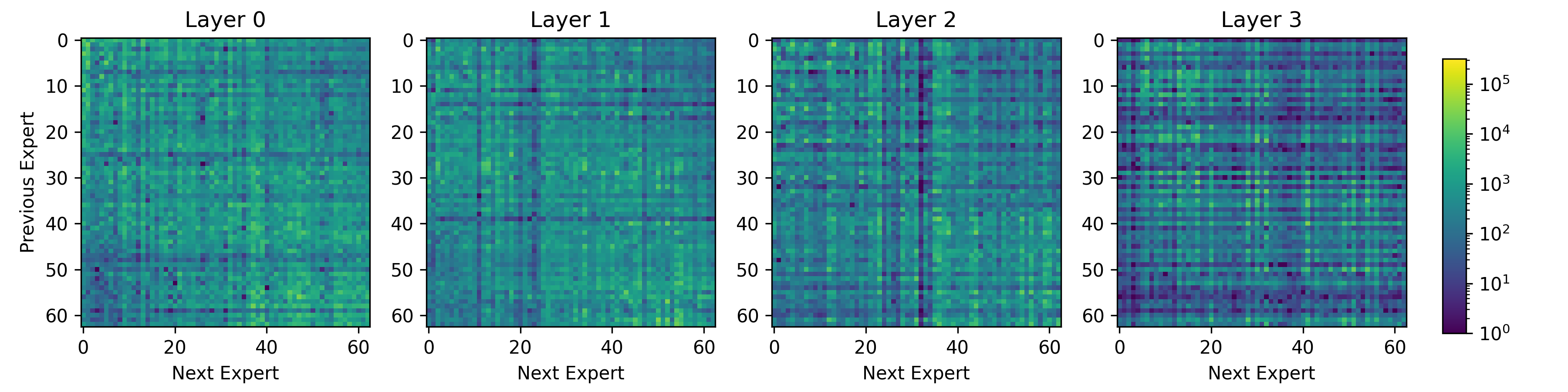}
    \caption{Layer-wise routing pattern on SlimPajama under the 4-experts-per-iteration, 2-iteration setup (2k training steps).}
    \label{fig:routing_4tpk2itr_slimpajama_2k}
\end{figure}

\begin{figure}[H]
    \centering
    \includegraphics[width=\linewidth]{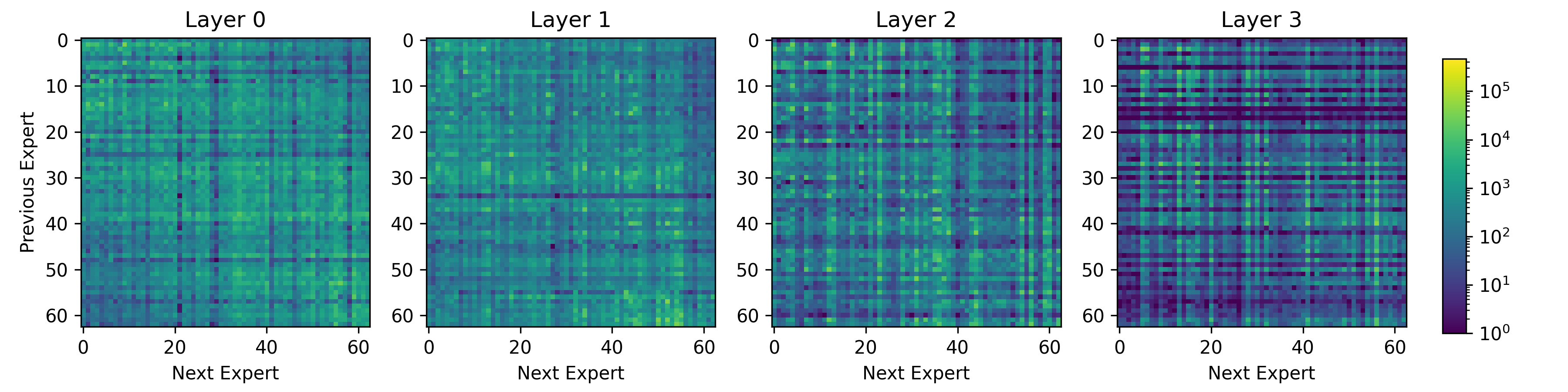}
    \caption{Layer-wise routing pattern on SlimPajama under the 4-experts-per-iteration, 2-iteration setup (10k training steps).}
    \label{fig:routing_4tpk2itr_slimpajama_10k}
\end{figure}

\begin{figure}[H]
    \centering
    \includegraphics[width=\linewidth]{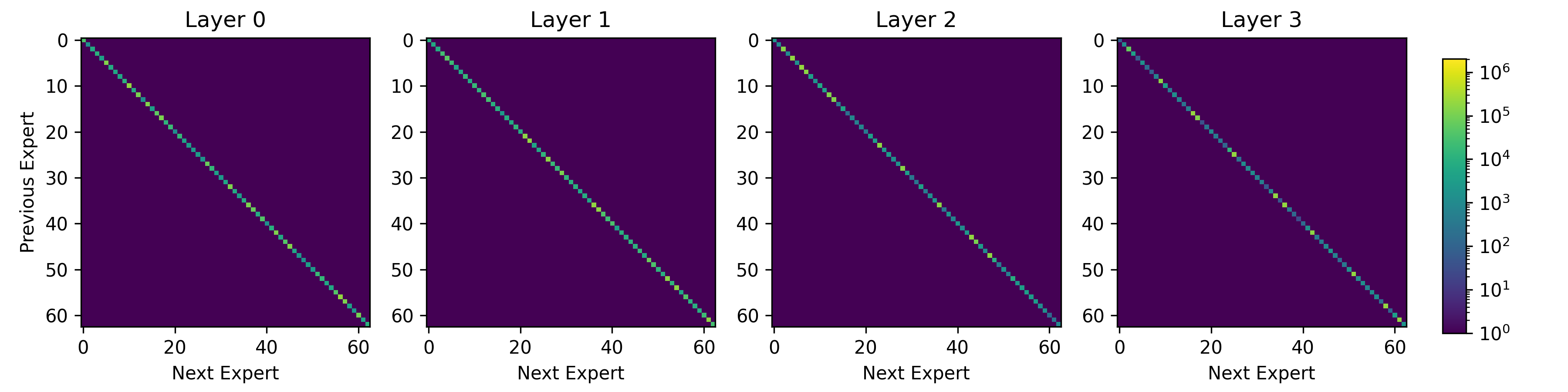}
    \caption{Layer-wise routing pattern on MetamathQA under the 8-experts-per-iteration, 1-iteration setup (2k training steps).}
    \label{fig:routing_8tpk1itr_metamathqa_2k}
\end{figure}

\begin{figure}[H]
    \centering
    \includegraphics[width=\linewidth]{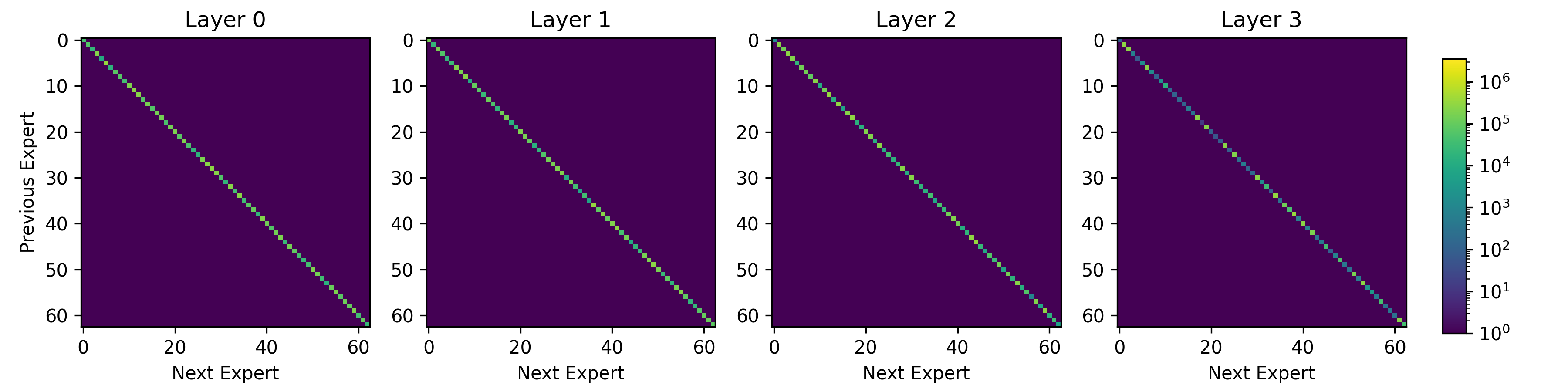}
    \caption{Layer-wise routing pattern on SlimPajama under the 8-experts-per-iteration, 1-iteration setup (2k training steps).}
    \label{fig:routing_8tpk1itr_slimpajama_2k}
\end{figure}

\begin{figure}[H]
    \centering
    \includegraphics[width=\linewidth]{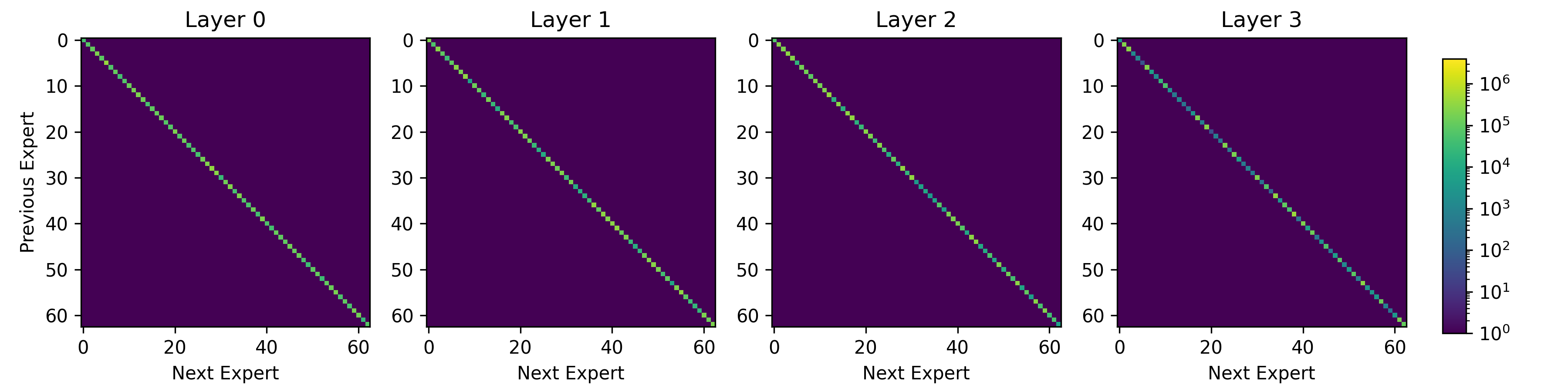}
    \caption{Layer-wise routing pattern on SlimPajama under the 8-experts-per-iteration, 1-iteration setup (10k training steps).}
    \label{fig:routing_8tpk1itr_slimpajama_10k}
\end{figure}

Each visualization in Figures~\ref{fig:routing_4tpk2itr_metamathqa_2k}--\ref{fig:routing_8tpk1itr_slimpajama_10k} consists of four heatmaps, corresponding to the four transformer layers in our model. These heatmaps characterize expert transition behaviors across CoE iterations. Specifically, each heatmap presents a two-dimensional expert-to-expert transition matrix, where the x-axis denotes the \textit{Next Expert} (i.e., experts assigned in iteration $t+1$), and the y-axis represents the \textit{Previous Expert} (i.e., experts assigned in iteration $t$). The intensity of each cell reflects the normalized frequency that tokens, previously processed by a given expert, are routed to another expert in the subsequent iteration at the same layer.

\subsection{Effect of Dataset on Expert Co-Activation Patterns}

We observe a consistent trend across datasets: \textbf{expert transitions in general-domain corpora (SlimPajama) are more evenly distributed}, whereas transitions in domain-specific data (MetaMathQA) tend to be more concentrated (\textit{e.g.}, Figure~\ref{fig:routing_8tpk2itr_slimpajama_2k} vs. Figure~\ref{fig:routing_8tpk2itr_metamathqa_2k}). It suggests that \textbf{experts in general-domain settings engage in a richer and more diverse interaction pattern}, potentially due to the broader range of linguistic phenomena. Interestingly, we also find that both SlimPajama and MetaMathQA exhibits \textbf{increasing expert specialization in the final layers compared to the early layers}, as evidenced by the emergence of dominant transition paths in the last one or two layers.

\subsection{Temporal Dynamics of Expert Co-Activation During Training}

\begin{figure}[H]
    \centering
    \includegraphics[width=\linewidth]{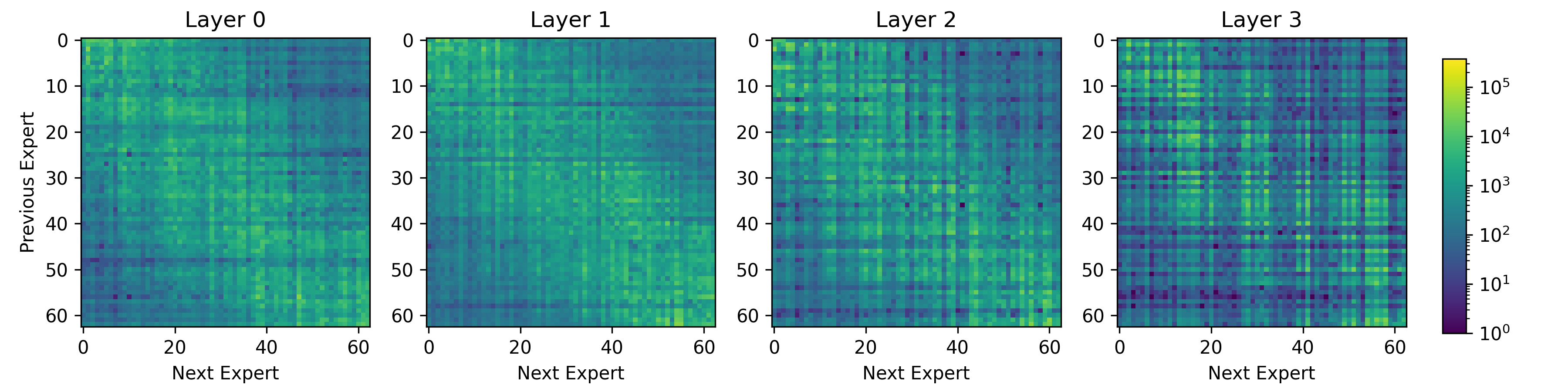}
    \caption{Layer-wise routing pattern on SlimPajama under the 8-experts-per-iteration, 2-iteration setup (2k training steps).}
    \label{fig:routing_8tpk2itr_slimpajama_2k}
\end{figure}

\begin{figure}[H]
    \centering
    \includegraphics[width=\linewidth]{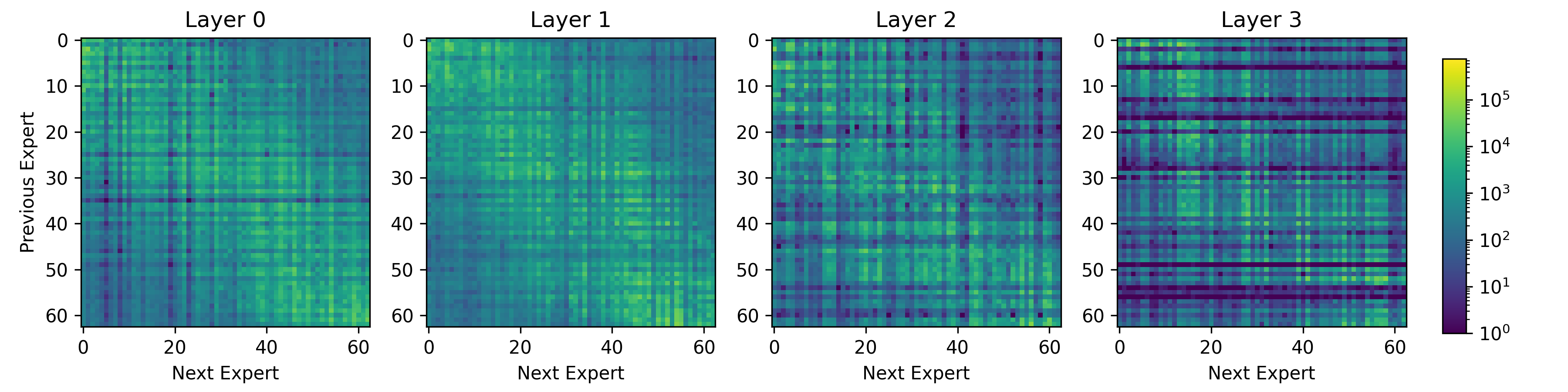}
    \caption{Layer-wise routing pattern on SlimPajama under the 8-experts-per-iteration, 2-iteration setup (10k training steps).}
    \label{fig:routing_8tpk2itr_slimpajama_10k}
\end{figure}

On the general-domain dataset SlimPajama, we observe that expert transitions become progressively more concentrated as training proceeds—particularly in deeper layers such as layer 3 (\textit{cf.}~Figure~\ref{fig:routing_8tpk2itr_slimpajama_2k} vs.~Figure~\ref{fig:routing_8tpk2itr_slimpajama_10k}). As the model is exposed to more diverse textual patterns, it gradually converges on a smaller subset of routing pathways that are reused consistently. This is reflected visually in the heatmaps: \textbf{bright regions grow brighter, and dark regions become darker}, indicating increasing polarization in expert assignment. Such a trend suggests that the model is identifying persistent and efficient expert transitions that generalize across large-scale linguistic contexts. In effect, \textbf{the routing structure simplifies over time}—emphasizing a few highly specialized paths that dominate token flow in later layers.

\begin{figure}[H]
    \centering
    \includegraphics[width=\linewidth]{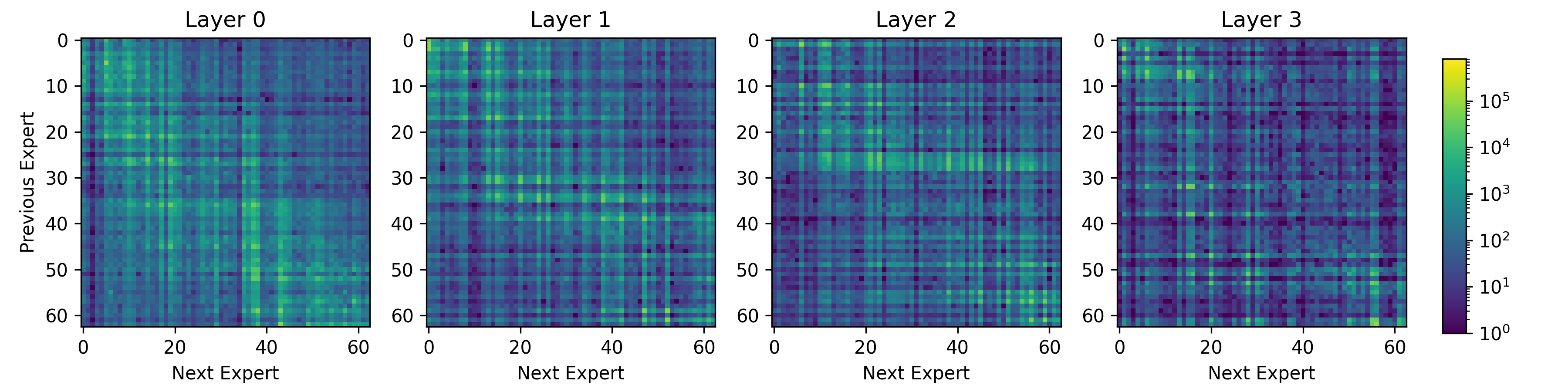}
    \caption{Layer-wise routing pattern on MetamathQA under the 8-experts-per-iteration, 2-iteration setup (2k training steps).}
    \label{fig:routing_8tpk2itr_metamathqa_2k}
\end{figure}

\begin{figure}[H]
    \centering
    \includegraphics[width=\linewidth]{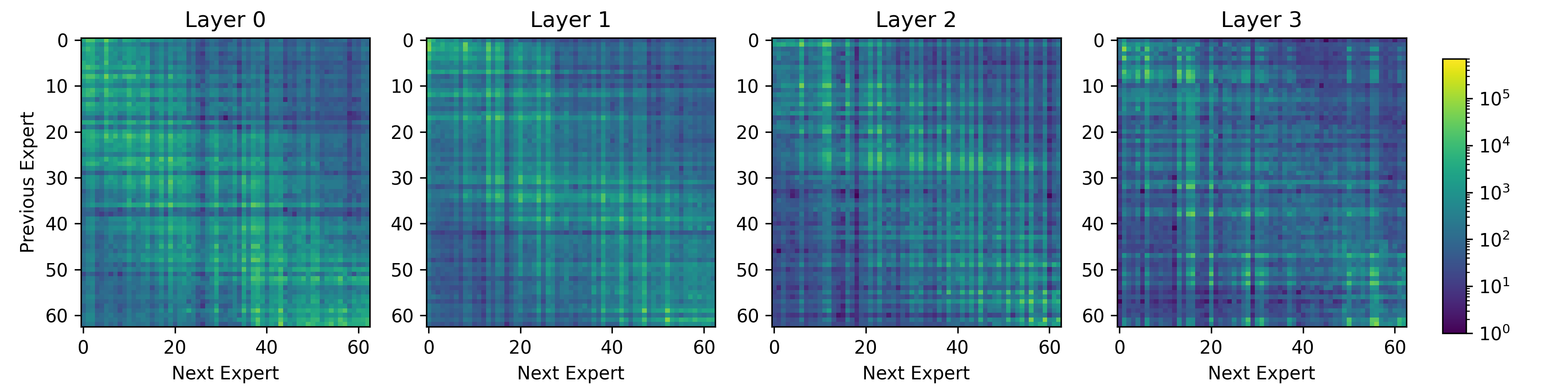}
    \caption{Layer-wise routing pattern on MetamathQA under the 8-experts-per-iteration, 2-iteration setup (10k training steps).}
    \label{fig:routing_8tpk2itr_metamathqa_10k}
\end{figure}

In contrast, on the more focused domain of MetaMathQA, expert transitions begin with a relatively narrow distribution—most tokens are routed similarly across iterations, resulting in sharp, localized activation patterns in early training (\textit{cf.}~Figure~\ref{fig:routing_8tpk2itr_metamathqa_2k}). However, as training progresses, these patterns become more diffuse (\textit{cf.}~Figure~\ref{fig:routing_8tpk2itr_metamathqa_10k}), reflecting \textbf{a growing diversity in how tokens are routed across experts}. This divergence likely arises because the model initially relies on uniform strategies when the domain is small and predictable, but later discovers the benefits of assigning different reasoning problems to different expert transitions. That is, \textbf{instead of collapsing to fixed pipelines, the model in MetaMathQA explores increasingly differentiated expert flows} as it learns to segment mathematical tasks by latent structure.

\subsection{Intra-Layer Expert Flow and Role Differentiation}

\begin{figure}[H]
    \centering
    \includegraphics[width=\linewidth]{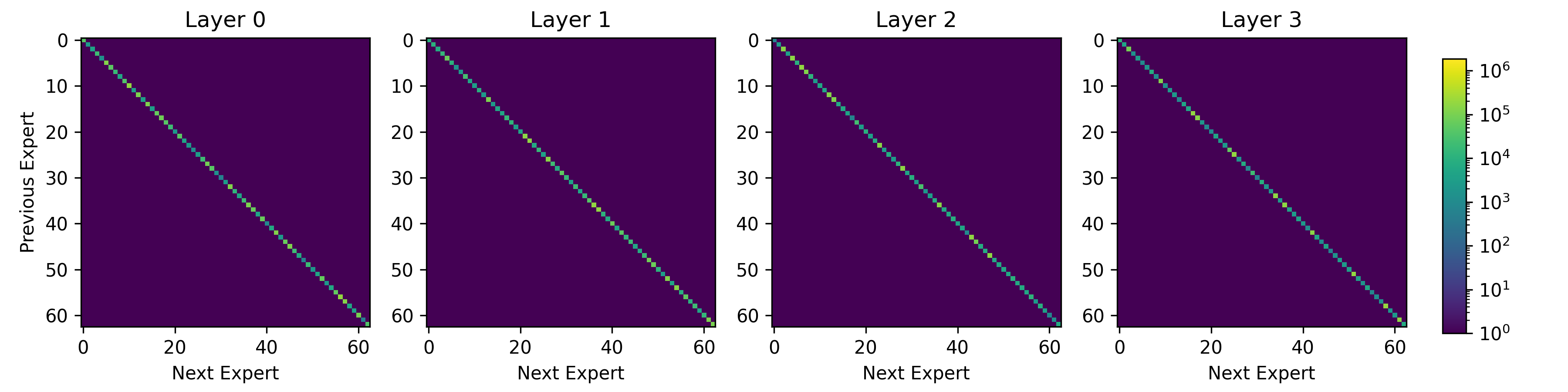}
    \caption{Layer-wise routing pattern on MetamathQA under the 8-experts-per-iteration, 1-iteration setup (10k training steps).}
    \label{fig:routing_8tpk1itr_metamathqa_10k}
\end{figure}

Within each layer, the diagonal intensity remains relatively low, indicating that \textbf{experts tend not to reprocess tokens they previously handled}. It affirms the \textbf{flowing nature of CoE}, where tokens are progressively transformed by distinct experts. However, we do observe a clear asymmetry between rows and columns in some matrices. Specifically, \textbf{certain experts frequently act as entry points}—handling more tokens in earlier iterations (brighter rows), while others serve as \textbf{accumulators}—processing aggregated representations in later iterations (brighter columns). This role differentiation is particularly pronounced on MetaMathQA (see Figure~\ref{fig:routing_8tpk1itr_metamathqa_10k}), suggesting emergent task-driven specialization among experts.

\subsection{Comparison with MoE: Stability of Expert Usage}

Unlike MoE, which often suffers from expert dropout or collapse in deeper layers during late training stages, CoE maintains robust expert utilization. Specifically, we find that in CoE, \textbf{each expert tends to be predominantly used in either the first or second layer}, but rarely vanishes entirely. This distribution alleviates the expert  collapse issue observed in MoE~\cite{fedus2022switch}, i.e., the CoE routing scheme could \textbf{naturally induce expert usage diversity and layer-specific specialization}.

\vspace{-5pt}
\vspace{-5pt}
\vspace{-5pt}
\vspace{-5pt}

\end{document}